\pdfoutput=1
\documentclass[11pt]{article}

\usepackage[]{ACL2023}
\usepackage{times}
\usepackage{latexsym}
\usepackage[T1]{fontenc}
\usepackage{hyperref}
\usepackage{url}
\usepackage{graphicx}
\usepackage{wrapfig,lipsum,booktabs}
\usepackage{caption}
\usepackage{multirow}

\usepackage{diagbox}
\usepackage{colortbl}
\usepackage{longtable}

\usepackage[utf8]{inputenc}
\usepackage[most]{tcolorbox}

\usepackage{microtype}

\usepackage{inconsolata}

\usepackage{amsmath,amsfonts,bm}

\def\eqref#1{equation~\ref{#1}}

\def\1{\bm{1}}

\DeclareMathAlphabet{\mathsfit}{\encodingdefault}{\sfdefault}{m}{sl}
\SetMathAlphabet{\mathsfit}{bold}{\encodingdefault}{\sfdefault}{bx}{n}

\usepackage{hyperref}
\usepackage{url}
\usepackage{graphicx}
\usepackage{wrapfig,lipsum,booktabs}
\usepackage{caption}
\usepackage{multirow}

\usepackage{colortbl}
\usepackage{tabularx,booktabs}

\definecolor{easy}{RGB}{83, 255, 191}
\definecolor{mid}{RGB}{233, 255, 83}
\definecolor{hard}{RGB}{255, 83, 147}
\definecolor{sA}{RGB}{173,216,230}
\definecolor{sB}{RGB}{144,238,144}
\definecolor{cA1}{RGB}{255,182,193}
\definecolor{cA2}{RGB}{255,160,122}
\definecolor{cB1}{RGB}{221,160,221}
\definecolor{cB2}{RGB}{218,112,214}

\usepackage{algpseudocode}
\usepackage{framed}
\usepackage{quoting}
 \usepackage{float} 
  \usepackage{threeparttable} 

\usepackage[most]{tcolorbox}
\usepackage{lipsum}
\newenvironment{shadedquotation-1}
 {\colorlet{shadecolor}{blue!15} 
\begin{shaded*}
\begin{quoting}
\item
 }
 {\end{quoting}
\end{quoting}
}

\newenvironment{shadedquotation-2}
 {\colorlet{shadecolor}{orange!15} 
 \begin{shaded*}
\begin{quoting}
\item
 }
 {\end{quoting}
\end{shaded*}
}

\newenvironment{shadedquotation-3}
 {\colorlet{shadecolor}{gray!7} 
 \begin{shaded*}
\begin{quoting}
\item
 }
 {\end{quoting}
\end{shaded*}
}

\title{Language Models Don't Learn the Physical Manifestation of Language}

\author{Bruce W. Lee \\
  University of Pennsylvania\\
  \texttt{brucelws@seas.upenn.edu} \\\And
  JaeHyuk Lim \\
  University of Pennsylvania\\
  \texttt{jaehyuk@sas.upenn.edu} \\}
  
\begin{document}

\maketitle

    \begin{abstract}

We argue that language-only models don't learn the physical manifestation of language. 
We present an empirical investigation of visual-auditory properties of language through a series of tasks, termed \textsc{H-Test}.
These tasks highlight a fundamental gap between human linguistic understanding and the sensory-deprived linguistic understanding of LLMs. 
In support of our hypothesis, \textbf{1. }deliberate reasoning (Chain-of-Thought), \textbf{2. }few-shot examples, or \textbf{3. }stronger LLM from the same model family (LLaMA 2 13B $\rightarrow$ LLaMA 2 70B) has no significant effect on \textsc{H-Test} performance. 

We bring in the philosophical case of Mary, who learns about the world in a sensory-deprived environment as a useful conceptual framework to understand how language-only models learn about the world \citep{jackson1986mary}.
Our experiments show that some of the strongest proprietary LLMs stay near random chance baseline accuracy of 50\%, highlighting the limitations of linguistic knowledge acquired in the absence of sensory experience. 
Our code and data are available at <github.com/brucewlee/h-test>.

\end{abstract}
    \section{Introduction}
The advent of large language models (LLMs) like ChatGPT has revolutionized the field of computational linguistics, offering unprecedented insights into the mechanics of language processing and generation. 
Yet, a fundamental question persists: How does a language model ``experience'' and consequently learn language, and how does this differ from human language acquisition? 
Our aim lies in the systematic exploration of these perceptual-level differences, focusing on the distinction between the acquisition of language through textual data alone versus the multimodal sensory experiences inherent to human learning.

\begin{figure}[t]
\centering
\includegraphics[width=0.5\textwidth]{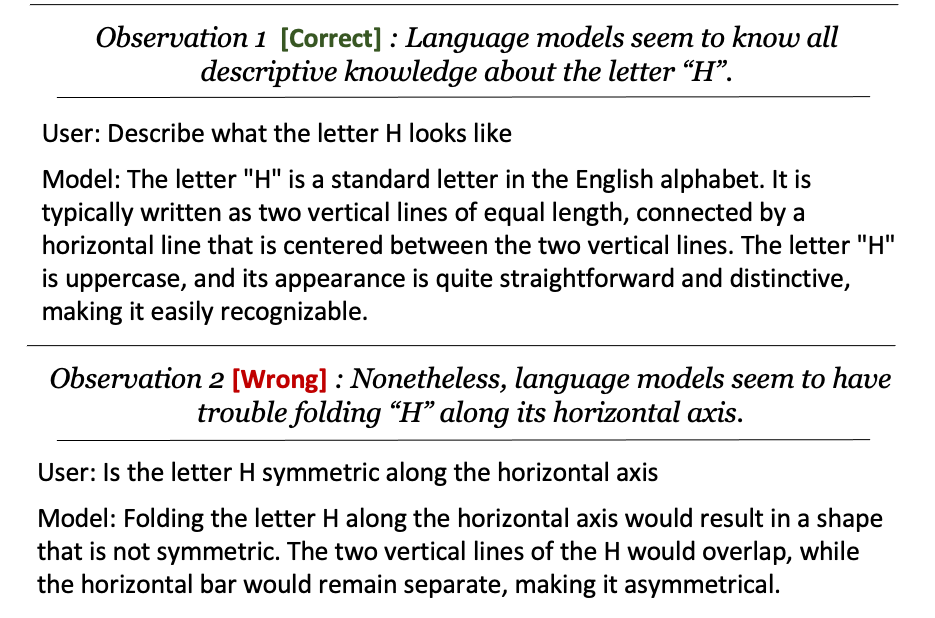}
\captionof{figure}{\textbf{Conundrum}: What information is fundamentally absent in the current language training dynamics?}
\label{fig:observations}
\vspace{-4mm}
\end{figure}

Humans experience language through a rich sensory interplay involving visual, auditory, and tactile stimuli \citep{lieberman2002nature}, integrating these sensory inputs to form a comprehensive understanding of language. In contrast, LLMs are trained on vast corpora of text data, quite devoid of sensory context. 
They process language as a series of tokens, learning patterns and associations between these tokens without any direct experience of the concepts they represent. 
This sensory deprivation raises critical questions about the completeness and depth of the ``understanding'' of language that LLMs can achieve through text-based learning alone.
Roughly speaking, LLM ``perceives without sensing''.

To probe these questions, we introduce a series of tasks, collectively referred to as the \textsc{H-Test}, inspired by philosophical thought experiments.
\textsc{H-Test} is designed to assess the ability of LLMs to learn the physical manifestation of language -- that is, how our language looks in our eyes.
These tasks challenge the models to demonstrate a form of language understanding that is closer to human cognitive processes.

This exploration is motivated by the philosophical debate on the nature of knowledge and understanding, particularly the thought experiment involving Mary, a scientist who knows everything about the color red but has never experienced it directly. 
By analogy, can an LLM that has never ``seen'' or ``heard'' truly understand the full linguistic knowledge that derives its significance from visual and auditory experiences?

\begin{itemize}
\item \textbf{Research Aim 1}: To identify and categorize the types of linguistic information that may remain perpetually out of reach for LLMs trained exclusively on text data, thereby highlighting potential linguistic ``blind spots'' \citep{zimmerman2023blind} in their learning.
\item \textbf{Research Aim 2}: To examine the extent to which a language model trained solely on text can achieve a meaningful level of visuospatial ability. 
Is it possible for these models to conceptualize and employ visual imagery (like F + \_ $\rightarrow$ E) despite their training limitations? 
Researchers in neuroscience and psychology have asked similar research questions on congenitally blind people, and they report rather mixed results \citep{likova2018transfer, ruggiero2010role, vanlierde2004abilities, aleman2001visual}.
\end{itemize}

Figure \ref{fig:observations} encapsulates the central philosophical conundrum that guides our inquiry.
While there are a number of anecdotal examples of state-of-the-art LLMs struggling with understanding and utilizing these orthographic components of language (\textit{ChatGPT doesn't know if ``r'' is in the word Blueberry} in Appendix \ref{App:anecdotal}), we quantify and show that this deficiency is rather difficult to solve by commonly pursued LLM research directions. Adding more orthography data will not trivially solve this issue.

\begin{table*}[ht]
\centering
\footnotesize
\begin{tabular}{lp{1.5in}p{1.5in}p{1.5in}}
\toprule
\textbf{Task} &\textbf{Criteria} & \textbf{Group A (Example)} & \textbf{Group B (Example)}\\ 
\midrule
Uppercase & Group A has one random letter in uppercase. & owl calculates in \textcolor{blue}{\textbf{T}}he hall adventurously. & iguana reads with interest expertly. \\ 
\midrule
Starts Vowel & Group A always starts with a vowel. & \textcolor{blue}{\textbf{E}}ngineer sings with interest eagerly. & \textcolor{red}{\textbf{L}}ion cooks delicious meals enthusiastically. \\ 
\midrule
End Punctuation & Group A always ends with a punctuation.& Octopus in Space Butterfly\textcolor{blue}{\textbf{...}} & Astronaut \textcolor{red}{\textbf{...}} through the telescope Elephant \\ 
\midrule
Palindrome & Group A reads the same forward and backward. & \textcolor{blue}{\textbf{wow}} & Debby \\ 
\midrule
End Ly & Group A always ends with ``ly''. & Tree reads in the studio adventurous\textcolor{blue}{\textbf{ly}}. & Cat teaches with interest enough. \\ 
\midrule
Spelled Math & Group A contains no spelled-out math notation. & The cube of equals to 3. & The square root \textcolor{red}{\textbf{+}} 4. \\ 
\midrule
Spelled Number & Group A contains spelled-out number. & River talks \textcolor{blue}{\textbf{four}} computers intently. & Butterfly ticks \textcolor{red}{\textbf{3}} apples intently. \\ 
\midrule
Rhyme & Group A words always rhyme with each other.& \textcolor{blue}{\textbf{get fat}} & ask go \\ 
\midrule
Repeated Word & Group A repeats one random repeated word. & Book wall \textcolor{blue}{\textbf{fast fast}} jumps. & intently drives meals Car. \\ 
\midrule
Hyphenated Word & Groups A contains one hyphenated expression. & A friend reads on the wall in a \textcolor{blue}{\textbf{up-to-date}} manner. & A car ticks with passion fast. \\ 
\bottomrule
\end{tabular}

\caption{\textbf{A/B Classification:} Ten tasks that are easy if one reads language (senses before perceiving) but difficult without senses. Bold parts show the tell-tale components that easily reveal the classification criterion between A and B. While ensuring that both group A and B instances are structurally the same, we intentionally use grammatically wrong sentences for both groups in some tasks to prevent LLMs from classifying based on linguistic correctness. To show these tasks are rather obvious to human readers, we give exemplars in Appendix \ref{App:abExemplar}.}
\label{tab:tasks}
\vspace{-4mm}
\end{table*}

\section{What Is It Like to Be a ChatGPT?}
In the realm of LLMs, a predominant belief is that scaling up data, model size, and computational power invariably leads to enhanced performance \citep{openai2023gpt4, rae2021scaling}, sometimes exceeding expectations \citep{bubeck2023sparks, wei2022emergent, wei2022inverse}. 
The recent report on the emergence of world representations within LLMs further emphasizes this point \citep{li2023emergent, nanda_othello_2023}, suggesting a diminishing return on the effort to identify tasks that cannot be addressed by scaling.

An intriguing perspective on this topic, as proposed by \citet{zimmerman2023blind}, is the idea of comprehending the ``experience'' of a ChatGPT-like model. 
This line of thought is crucial, as it highlights a common oversimplification: the tendency to describe LLMs as being trained in ``text'' or ``language,'' as perceived by humans. 
This tendency to describe things as they seem to us is also discussed as anthropomorphic behavior \citep{shanahan2023talking}.
In defense of scientific accuracy, \citet{zimmerman2023blind} advances this discussion by distinguishing between two types of information contained within language: Diegetic Information, which encompasses the internal, semantic, and propositional aspects of language, and Supradiegetic Information, which refers to the physical form of language, such as the shape of letters and sounds of syllables.

Our interpretation aligns with this framework. Diegetic information in training data roughly corresponds to the semantics of a language. 
The efficacy of language modeling in teaching this aspect is evident from the consistent improvement in tasks like knowledge-based question answering and other semantics-heavy applications \citep{biderman2023pythia, chowdhery2022palm}. 
However, the supradiegetic aspect, which involves the sensory perception of language, remains a less explored territory for language modeling. 
This is evident from the fragmental evidence of language models' struggles in downstream tasks with heavy stylistic features, such as automated essay scoring \citep{mizumoto2023exploring, uto2020neural} and readability assessment \citep{hou2022promoting, lee2021pushing}.

To anchor our discussion, we reference the thought experiment by \citet{jackson1986mary}, featuring ``Mary,'' a scientist confined to a black and white room, learning about the world through monochromatic means. 
Despite her knowledge of the descriptive aspects of human color vision, her understanding is incomplete until she ``sees'' color firsthand. 

\begin{enumerate}
\item Mary possesses complete descriptive knowledge about human color vision prior to her release.
\item Yet, upon seeing color directly, Mary will likely acquire new knowledge.
\item Thus, certain knowledge is inherently sensory-dependent.
\end{enumerate}

Drawing parallels to our investigation, we posit that while LLMs like ChatGPT are adept at processing the diegetic content of language, their understanding of supradiegetic elements remains rudimentary. 
The fundamental difference in how humans and machines experience language inevitably leads to some level of incompleteness in learned linguistic knowledge \citep{driess2023palm}, and therefore, will not be able to solve certain language tasks, inherently.

    \section{\textsc{H-Test}}
Do language models truly comprehend what the letter ``H'' looks like (Figure \ref{fig:observations})?
This inquiry goes beyond the basic propositional knowledge and delves into the LLMs' capacity for visual representation and comprehension.

As illustrated in Figure \ref{fig:observations}, advanced LLMs like ChatGPT demonstrate a competent understanding of how the letter ``H'' is described. 
This capability, however, does not necessarily equate to a comprehensive understanding of how the letter ``H'' looks like, at least in the philosophical sense. 
This is reminiscent of Mary's incomplete knowledge in \citet{jackson1986mary}'s thought experiment. 
Mary's knowledge of color, although extensive in a physical and descriptive sense, was fundamentally incomplete until Mary saw color directly. 
Similarly, visual elements, such as the appearance of letters, or auditory elements, such as the rhyming sounds of two words, may be inherently limited, or very difficult to understand or use for LLMs.

\begin{table}[t]
\centering
\footnotesize
\resizebox{0.47\textwidth}{!}{%
\begin{tabular}{l|c|c|c|c|c|c}

\textbf{Task} & \rotatebox{90}{\textbf{J2 Ultra}} & \rotatebox{90}{\textbf{Claude V2}} & \rotatebox{90}{\textbf{LLaMA 2 70B}} & \rotatebox{90}{\textbf{Command-Li.}} & \rotatebox{90}{\textbf{GPT 3.5}} & \rotatebox{90}{\textbf{Luminous Sup.}} \\ 

\midrule

Uppercase        & 50.5 & 55.5 & 45.5 & 45.5 & 49.5 & 49.  \\ 
Start Vowel      & 51.  & 65.  & 49.  & 47.5 & 51.5 & 49.  \\ 
End Punctuation  & 48.  & 52.  & 47.5 & 54.5 & 52.5 & 52.  \\ 
Palindrome       & 51.  & 58.  & 40.  & 20.5 & 55.5 & 50.5 \\ 
End Ly           & 58.5 & 75.  & 46.  & 61.  & 46.5 & 51.  \\ 
Spelled Math     & 51.  & 69.5 & 45.5 & 40.5 & 63.5 & 50.5 \\ 
Spelled Number   & 51.  & 55.  & 45.  & 49.  & 49.5 & 49.5 \\ 
Rhyme            & 49.5 & 77.5 & 47.  & 51.  & 57.  & 51.5 \\ 
Repeated Word    & 49.5 & 53.  & 32.5 & 54.  & 50.5 & 51.  \\ 
Hyphenated Word  & 51.  & 58.  & 53.5 & 47.  & 46.5 & 50.  \\ 

\midrule
Average          & 51.1 & 60.4 & 44.8 & 47.1 & 52.3 & 50.4  \\

\bottomrule
\end{tabular}}
\caption{\textbf{State-of-the-art LLMs Struggle:} We report few-shot (at k = 50) accuracies on A/B classification tasks across six LLM services. We are reporting performances of the language-only models from AI21, Anthropic, Meta, Claude, OpenAI, and Aleph Alpha.}
\label{tab:proprietary}
\vspace{-6mm}
\end{table}

\begin{figure*}[t]
\centering
\includegraphics[width=\textwidth]{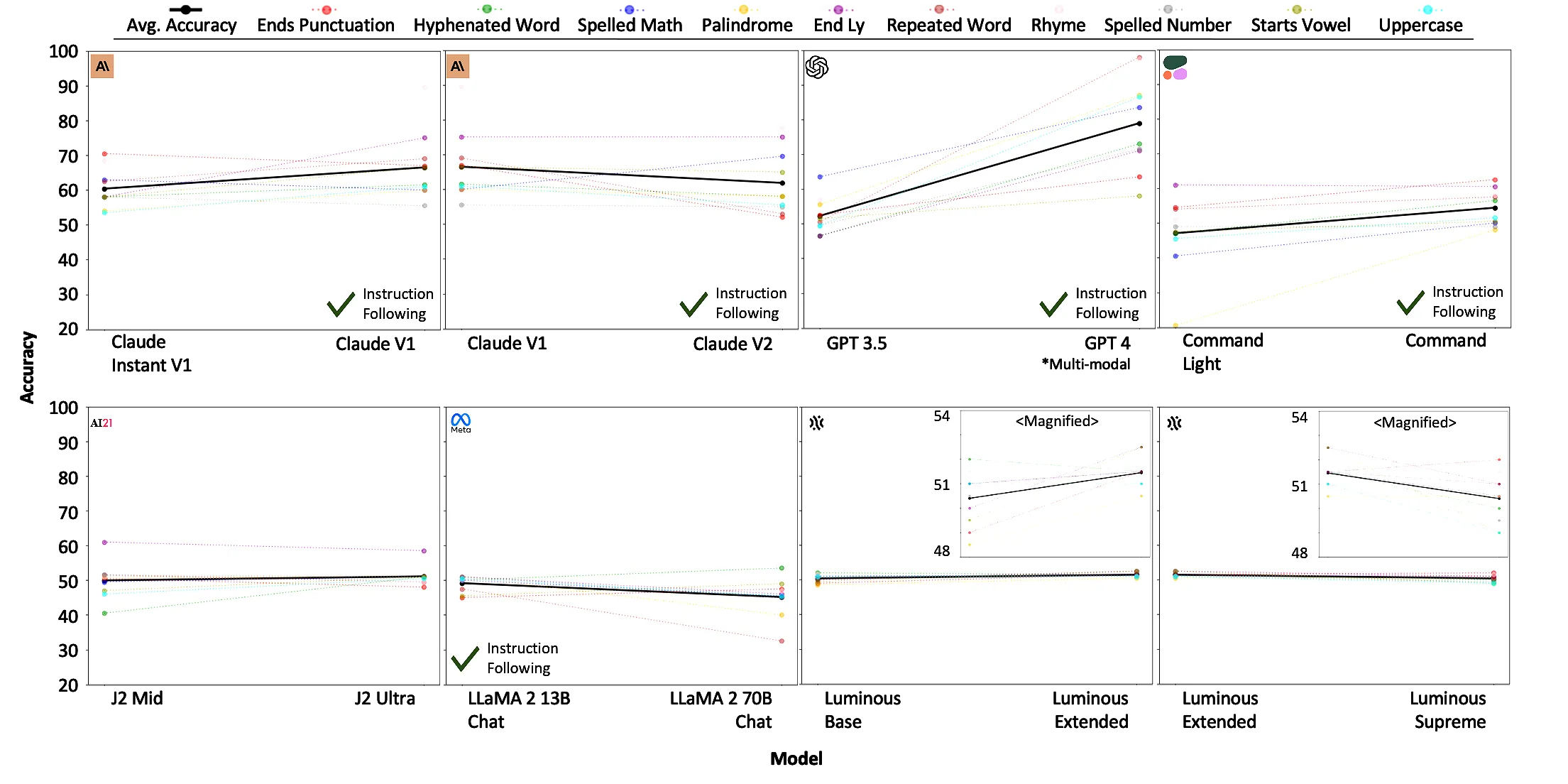}
\captionof{figure}{\textbf{Making Progress on Language-only Modeling Does Not Trivially Solve H-Test}: We test weaker models from the same family for models given in Table \ref{tab:proprietary} under the same few-shot (at k = 50) setup. The graphs for the Luminous model family are also shown in magnified versions to show that we are not depicting a flat line.}
\label{fig:intrafamily}
\vspace{-4mm}
\end{figure*}

To explore this hypothesis, we propose ten distinct classification tasks in Table \ref{tab:tasks} designed to test the limits of LLMs in processing the visual, auditory, and physical aspects of language that complement the semantic component. 
Solving these tasks requires a swift understanding and interlocking of how a language looks or sounds.
For us humans, such visual and auditory connections with language come naturally, but we argue that sensory-deprived LLMs find these tasks difficult to solve.

Each binary classification task has 200 balanced test instances (100 A, 100 B), along with 50 balanced few-shot instances (25 A, 25 B) that can be generated differently with varying random seeds.
In this paper, we use the fixed random seed of \texttt{12062023}, and other experimental setups are detailed in Appendix \ref{App:abDetail}.

\textbf{Precondition.} Given each test case and a set of example cases, the goal is to classify whether the test case belongs more to Group A or Group B.
Such a task formulation is also often referred to as few-shot prompting, where an LLM classifies the test cases from a number of pre-classified examples provided with the prompt.
Here, we use few-shot prompting instead of a direct inquisition like ``Does this sentence start with a vowel?'' 
Such prompts linguistically reveal the classification criterion, effectively making the task solvable by step-by-step sequencing of memorized facts like ``Elephant starts with E; E is a vowel; therefore, true.'' 

\textbf{Results.} Table \ref{tab:proprietary} reports \textsc{H-Test} accuracy on leading proprietary LLMs\footnote{We excluded GPT-4 from Table \ref{tab:proprietary} because GPT-4 is trained with multi-modality in mind \citep{openai2023gpt4}}.
Considering that the random chance accuracy on this binary classification task is 50\%, most tested models showed performance that is surprisingly close to this random chance baseline.
These models struggled to solve \textsc{H-Test} despite being given 50 examples.
This strongly suggests the confirmation of our initial conjecture that these sensory properties of language are ``blind spots'' in most language modeling formulations.
Next up, we show that \textsc{H-Test} is not trivially solvable through the commonly-pursued LLM research directions: 1. training a stronger, larger LLM with more data (which will probably train more orthographic descriptions of language), 2. serving more few-shot examples (which also serves more orthography-related data), 3. alternative prompting methods to induce multi-step reasoning.

\textbf{Observation 1. Insignificant Intra-Family Improvements.} 
If a stronger language-only system from the same model family, from the same lab, solves \textsc{H-Test} better, it would strongly hint that developing a stronger LLM can solve \textsc{H-Test} better.
In such a case, visual and auditory external features of language that we experience by sensing can be learned without actually ``sensing'' language, and our case will be negated.
In Figure \ref{fig:intrafamily}, we observe that a stronger model in the same model family often does not bring meaningful improvement on the \textsc{H-Test} performance. 
Some models, like Jurassic 2 and Luminous, showed almost neglectable average accuracy changes, tightly centered around the random baseline.
Though the details of most proprietary LLMs are not reported scientifically, the LLaMA 2 model family has rather thorough reports and shows that simply increasing the model size is not an apparent solution. 

Quite interestingly, GPT-4, the only model that was reported to have undergone multi-modal training \citep{openai2023gpt4}, made a particular improvement on \textsc{H-Test} compared to its precedent. However, this initial performance of the GPT-4 API couldn't be replicated when we accessed the API again a few weeks later. Instead, GPT-4o replicated a similar performance. Due to this inconsistency, the initial GPT-4 performance was removed from Table \ref{tab:full} and replaced with the GPT-4o performance.

In Section \ref{gpt4}, we make an attempt to explain this surprising improvement by comparing multi-modal and mix-of-experts open-source models. However, we fail to explain the improvement in GPT-4o level.

Though we are limited by the amount of knowledge on how GPT-4 was trained in a multi-modal way, such a result does hint at a good assumption that \textsc{H-Test} is indeed solvable when given a well-devised multi-modal system, or possibly, a large enough system \citep{wei2022inverse}. However, without a thorough report, no presumption about GPT-4 or GPT-4o model architecture can be made.

\begin{figure}[t]
\centering
\includegraphics[width=0.5\textwidth]{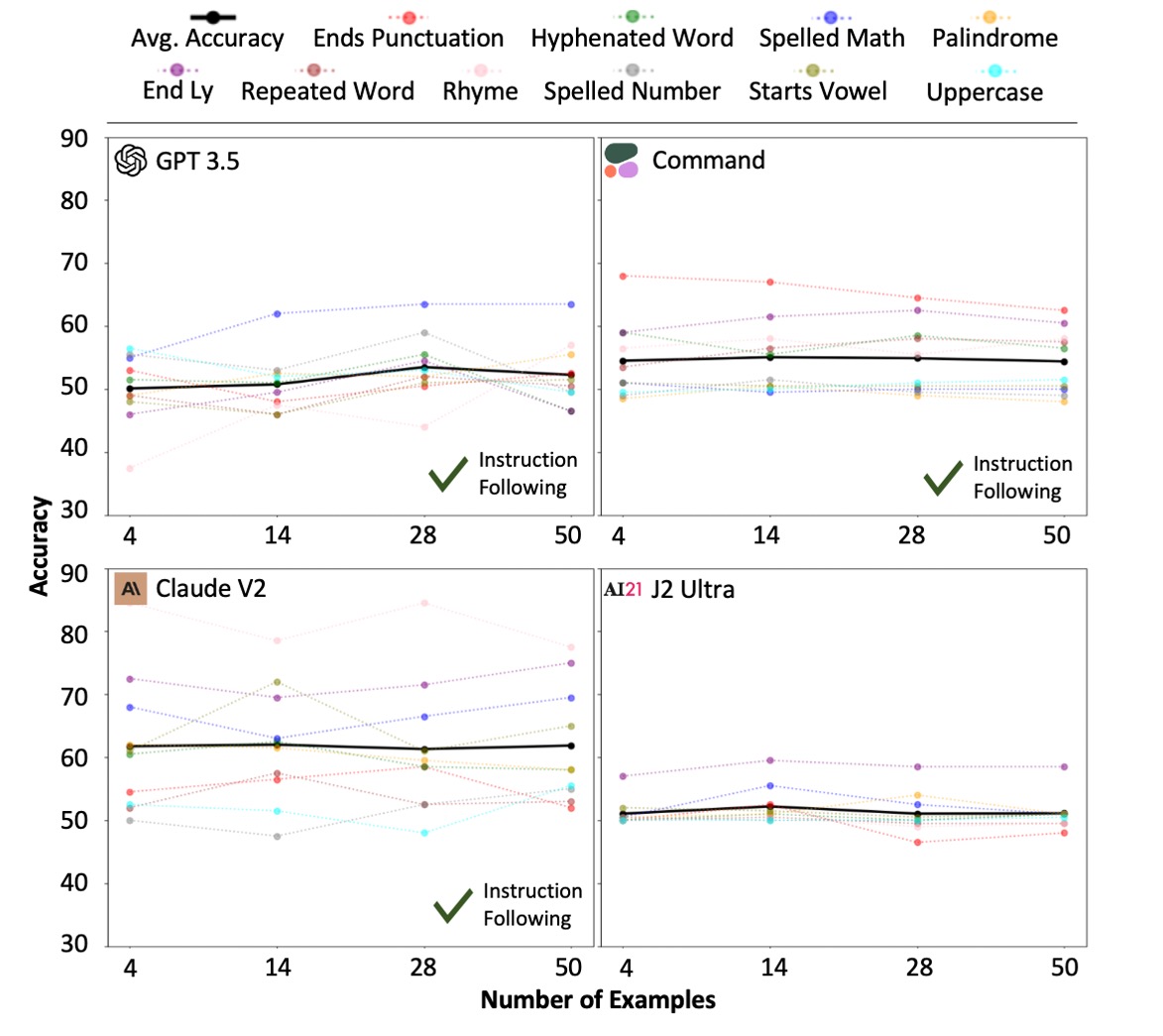}
\captionof{figure}{\textbf{Current LLMs Do Not Solve \textsc{H-Test} Better with More Examples}: We test four models from Figure \ref{fig:intrafamily} and test with different number of examples (k = \{4, 14, 28, 50\}). Though we acknowledge that subtask accuracy does vary at different few-shot setups, giving more or fewer examples does not significantly alter the \textsc{H-Test} performance on average. }
\label{fig:fewshot}
\vspace{-4mm}
\end{figure}

\begin{figure}[t]
\centering
\includegraphics[width=0.5\textwidth]{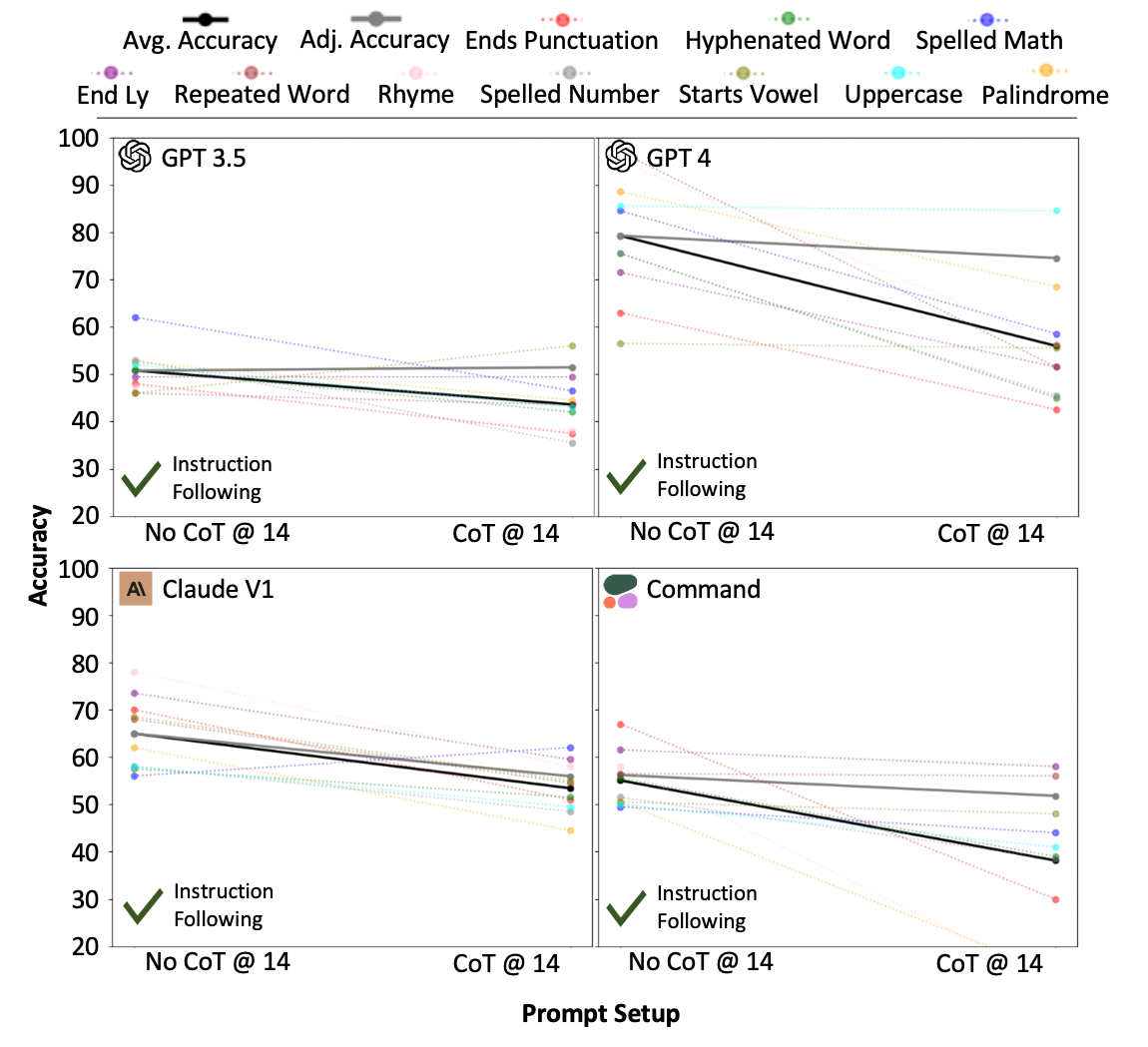}
\captionof{figure}{\textbf{\textsc{H-Test} is Not Meant to be Deliberately Reasoned}: We test four instruction-following and test with and without CoT prompt at k = 14. In general, we observe that CoT decreases performance. Adj. accuracy depicts the score, excluding the cases where the model did not generate a clearly interpretable CoT response. }
\label{fig:cotcomparison}
\vspace{-4mm}
\end{figure}

\textbf{Observation 2. Number of Examples Has Minimal Impact.} 
Few-shot prompting that we have been using to report the results in Table \ref{tab:proprietary} and Figure \ref{fig:intrafamily} is also commonly referred to as in-context learning (ICL) for its property to \textit{allow language models to learn tasks given only a few examples in the form of demonstration} \citep{dong2023survey, brown2020language}.
While ICL can often be as effective as fine-tuning \citep{duan2023exploring, NEURIPS2022_0cde695b}, certain studies report that language models internally and temporarily learn these examples to make inferences \citep{akyürek2023learning, dai2022can, min2022rethinking}. 
Combined with reports of cases where the increasing number of examples can improve performance \citep{duan2023exploring, chen2023demonstrations}, one would be able to make a convincing argument against our propositions if performance scaled with the number of examples. 
In such a case, the few-shot examples given in \textsc{H-Test} are indeed ``learnable.''

Figure \ref{fig:fewshot} reports respective performances with a varying number of examples. We only test stronger language-only models in light of the recent reports that ICL itself can be a learned ability \cite{wei2023larger, olsson2022incontext}. 
Even though the range of the few-shot setup tested is rather wide (from k = 4 to k = 50), the effect on the average accuracy is minimal.
That is, the number of examples given neither increases nor decreases performance significantly, strongly hinting that the LLM is simply not learning from \textsc{H-Test} examples cases. 
Such an observation strengthens our initial conjecture that the visual-auditory properties of language that we measure with \textsc{H-Test} are analogous to the ``blindspot'' of current language modeling formulations and will be difficult to learn without sensory experience of language.
Now, given that the few-shot examples given in \textsc{H-Test} are not ``learnable,'' is there an experimental setup that we deem is generally helpful for LLMs but rather misleads them in \textsc{H-Test}?

\begin{table*}[ht]
\centering
\footnotesize
\begin{tabular}{l|p{4.6in}}
\toprule
\textbf{Category} & \textbf{Prompt / Response on \textsc{H-Test}: Ends Punctuation} \\ 
\midrule
Few-Shot Examples  &   Input: Opera through the telescope School\textcolor{blue}{\textbf{.}} Label: A

Input: Octopus \textcolor{red}{\textbf{.}} along the valley Flower Label: B

Input: Elevator in space Lion\textcolor{blue}{\textbf{!}} Label: A

Input: Giraffe \textcolor{red}{\textbf{.}} all night Train Label: B

Input: Opera through the telescope Insect\textcolor{blue}{\textbf{.}} Label: A

Input: Eagle in space \textcolor{red}{\textbf{!}} Butterfly Label: B\\ 
\midrule
Test Case  &  Input: Kangaroo in the park Umbrella\textcolor{blue}{\textbf{.}} Label: \\ 
\midrule
CoT Response \textbf{[Wrong]} & 

In the given input sentence, \textcolor{red}{\textbf{the phrase ``in the park'' suggests a peaceful and natural setting, which might be more aligned with labels related to nature and tranquility. On the other hand, the word ``Umbrella'' brings to mind protection from the elements, which could be associated with indoor activities or certain weather conditions}}. Considering these factors, it is more likely that ``Kangaroo in the park Umbrella'' belongs to label B, which represents a description of animals in their natural habitats or surroundings.

Therefore, the final answer is: ``Kangaroo in the park Umbrella.'' \textcolor{red}{\textbf{Label: B}}. \\ 

\bottomrule
\end{tabular}
\caption{\textbf{Misleading CoT:} We report an example of a misleading CoT response generated from Command.}
\label{tab:mislead}
\vspace{-4mm}
\end{table*}

\textbf{Observation 3. Deliberate Reasoning Often Decreases Performance.} 
Lastly, we prompt LLMs to generate a Chain-of-Thought (CoT) before answering \citep{wei2022chain}. 
CoT is a widely explored prompting technique that is reported to consistently boost performance in a wide range of complex reasoning tasks, including mathematical, commonsense, symbolic, and logical reasoning \citep{chu2023survey}. 
With the development of Tree-of-Thought, \citet{yao2023tree} draws a particular connection to the ``System 2'' thinking of humans.

If LLMs benefit from such logical, step-by-step semantic reasoning on \textsc{H-Test}, this can also imply that \textsc{H-Test} is fundamentally solvable by developing stronger language-only models.
On the other hand, if CoT decreased or did not impact performance, this can strongly cue that \textsc{H-Test} is well-constructed to test our initial research target of simple, sensory-dependent experiential knowledge.

In Figure \ref{fig:cotcomparison}, we observe that CoT decreases performances in general, confirming our hypothesis.
GPT-4, which previously solved \textsc{H-Test} with over 80\% accuracy, also experienced a drop in performance, more strongly confirming our case. 
To better understand why CoT decreases performance, we analyze the CoT responses generated by the LLMs. 
Our findings reveal that while LLMs can produce coherent and logical chains of thought, these thought processes often do not align with the sensory aspects required to solve \textsc{H-Test} as shown in Table \ref{tab:mislead}. 
For instance, in tasks related to the visual representation of letters, LLMs tend to rely on abstract concepts, which do not translate into a genuine understanding of the visual or auditory characteristics. 
We discuss further experimental detail in Appendices \ref{App:abDetail}, \ref{App:responsesandopen}, and \ref{App:CoT}.

\textbf{Sanity Check: Human Performance.}
To validate whether the tasks within the \textsc{H-Test} are indeed as intuitive for humans as we hypothesized. Participants in the study were presented with the few-shot examples of the \textsc{H-Test} tasks without prior explanation or training. Participants were asked to identify the underlying pattern or rule that distinguished Group A from Group B examples.

As shown in Appendix \ref{App:HumanBaseline}, human participants achieved perfect scores on \textsc{H-Test}. In the ``Palindrome'' task, where one has to identify that Group A words read the same forward and backward, human accuracy was 100\%, reflecting our natural ability to visually recognize patterns. Similarly, in the auditory ``Rhyme'' task, participants could effortlessly identify rhyming words. We believe that this result reinforces the notion that human language comprehension is a multimodal process that current LLMs do not replicate.

\begin{table}[ht]
\centering
\footnotesize
\resizebox{0.47\textwidth}{!}{%
\begin{tabular}{l|p{1.3in}|l}
\toprule
\textbf{Type} & \textbf{Question} & \textbf{Options} \\ 
\midrule
Rotation&Given the letter "Z", what is it most likely to look like when rotated 90 degrees clockwise?  &   \underline{\textbf{N}}, A, D, T \\ 
\midrule
Rotation&Given the symbol "3", what does it most likely look like when rotated 180 degrees counterclockwise? &   f, d, \underline{\textbf{E}}, B \\ 
\midrule
Flipping&Given the letter "u", what is it most likely to look like when flipped along an imaginary vertical line in the middle? &   5, f, E, \underline{\textbf{u}} \\ 
\midrule
Add/Subtract&Given the letters "L" and "I", what is it most likely to look like when subtracted (L - I)? &   \underline{\textbf{\_}}, A, 0, J \\ 
\midrule
Composite&Given the letter "u", rotate it 180 degrees counter-clockwise and add it below the original letter. What is it most likely to look like? &   \underline{\textbf{H}}, c, J, Y \\ 

\bottomrule
\end{tabular}}
\caption{\textbf{Letter Geometry Examples:} Five examples of our letter geometry task with answers underlined.}
\label{tab:geometry}
\vspace{-4mm}
\end{table}

\begin{figure}[t]
\centering
\includegraphics[width=0.45\textwidth]{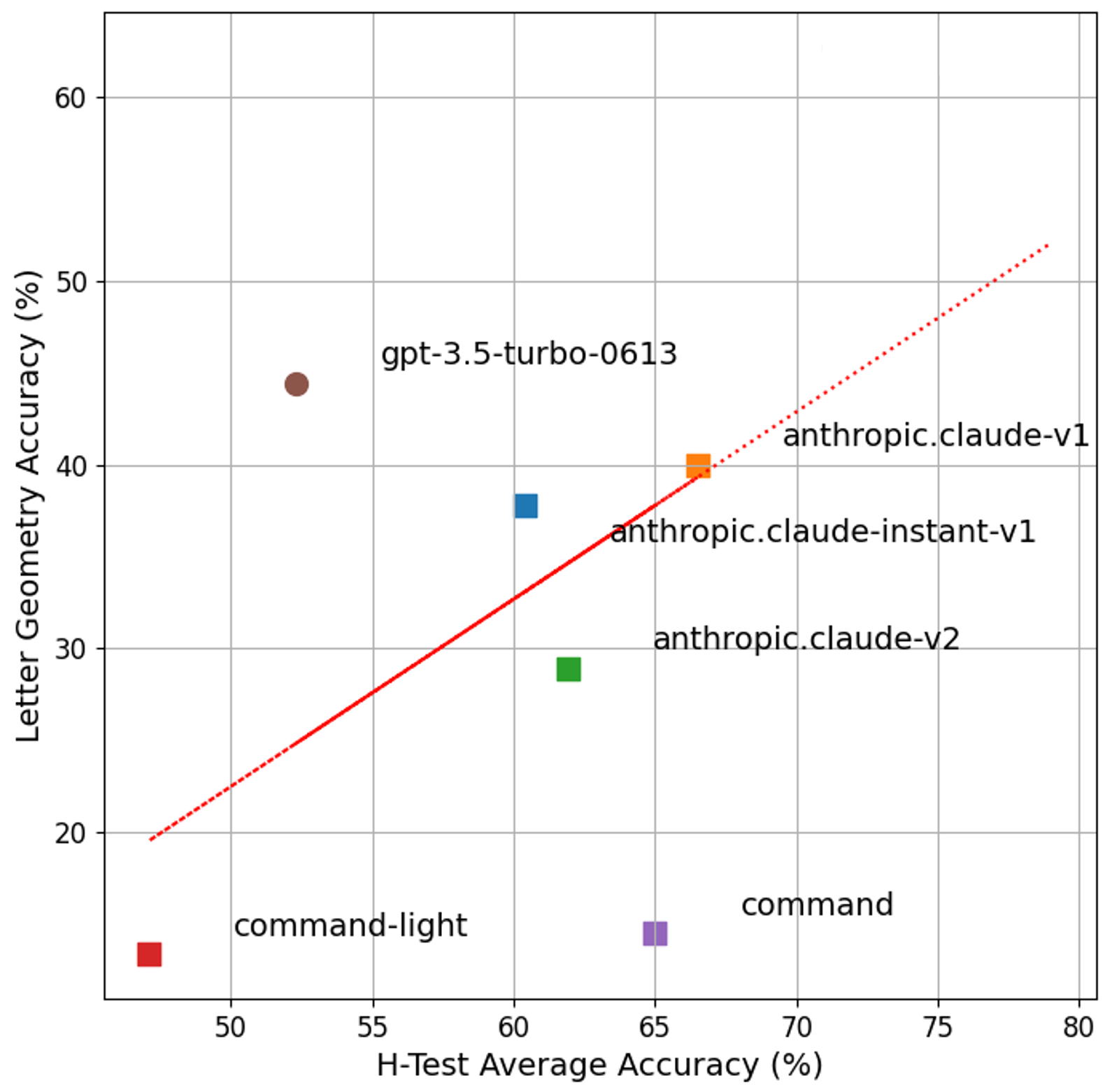}
\caption{\textbf{\textsc{H-test} vs. Letter Geometry:} We compare the accuracy of seven models on \textsc{H-Test} and Letter Geometry. The red line represents the linear best fit.}
\label{fig:htest_lettergeo}
\vspace{-4mm}
\end{figure}

\section{Reasoning Challenge: Letter Geometry}
So far, we have presented \textsc{H-Test} and showed that there are certain properties of language that are difficult to learn, as standard sensory-deprived language models do not go through the standard model of perception (stimulus $\rightarrow$ sense $\rightarrow$ perceive) in learning language. A particular and only exception was GPT-4 among the tested models. 

As a further challenge that tests not only the visual-auditory understanding of letters but also visuospatial reasoning abilities, we devise a more challenging task: Letter Geometry.
As shown in Table \ref{tab:geometry}, this task involves operations like flipping, rotation, addition, and subtraction on letters.
We do not include this as a standard part of the \textsc{H-Test} as performing well on this task likely involves step-by-step reasoning capabilities, which is not what we intended for \textsc{H-Test}.
However, we are still very interested in this letter geometry task as the smallest-sized token that a language model can process is often a letter \citep{elhage2021mathematical}, and the elements of the operations (e.g., `F' and `E') have no apparent semantic, phonetic, or logographic correlation that can be derived from the model’s training samples \citep{cheng2013chinese, lee2009korean}.
This multiple-choice task encourages the utilization of visual imagery (e.g., `F' + `\_' = `E') for above random baseline performance of 25\%.

The evaluation results are depicted in Figure \ref{fig:htest_lettergeo}. Most models remain close to a hypothetical baseline, struggling with the sensory-deprived nature of their training. This suggests that the capacity to process visuospatial information in the language is not simply a function of visual-auditory understanding of language but also of how models are trained to reason through sensory-like experiences.

We also assessed the performance improvement across different models from the same research labs. In the case of the Command models, the enhancement was marginal, at 1.11\%, which essentially translates to one additional correct response. A standout observation was with OpenAI's models: the shift from GPT-3.5 to GPT-4o marked a substantial increase of around 15\%.
    
\section{GPT-4o and Claude 3 Started Solving \textsc{H-Test} but How?}
\label{gpt4}

In Figure \ref{fig:intrafamily}, we have reported the seemingly unexplainable (jumping) performance improvement on \textsc{H-Test} from GPT-3.5 (language-only) to GPT-4 (multimodal). 
We observe a similar improvement from Claude 2 (language-only) to Claude 3 (multimodal) in Table \ref{tab:multimodal}.
This result is important as it shows that \textsc{H-Test} is indeed solvable (by a GPT-4-level system), but not through conventionally discussed language-only modeling techniques.

As further detailed in Table \ref{tab:full}, GPT-4o reaches human-level performance in palindrome and spelled math tasks, while all other tested models struggle at 50\% random baseline performance.
But considering that GPT-4o and Claude 3 are multi-modal, they lie beyond our initial claims. 

\begin{figure}[t]
\centering
\includegraphics[width=0.45\textwidth]{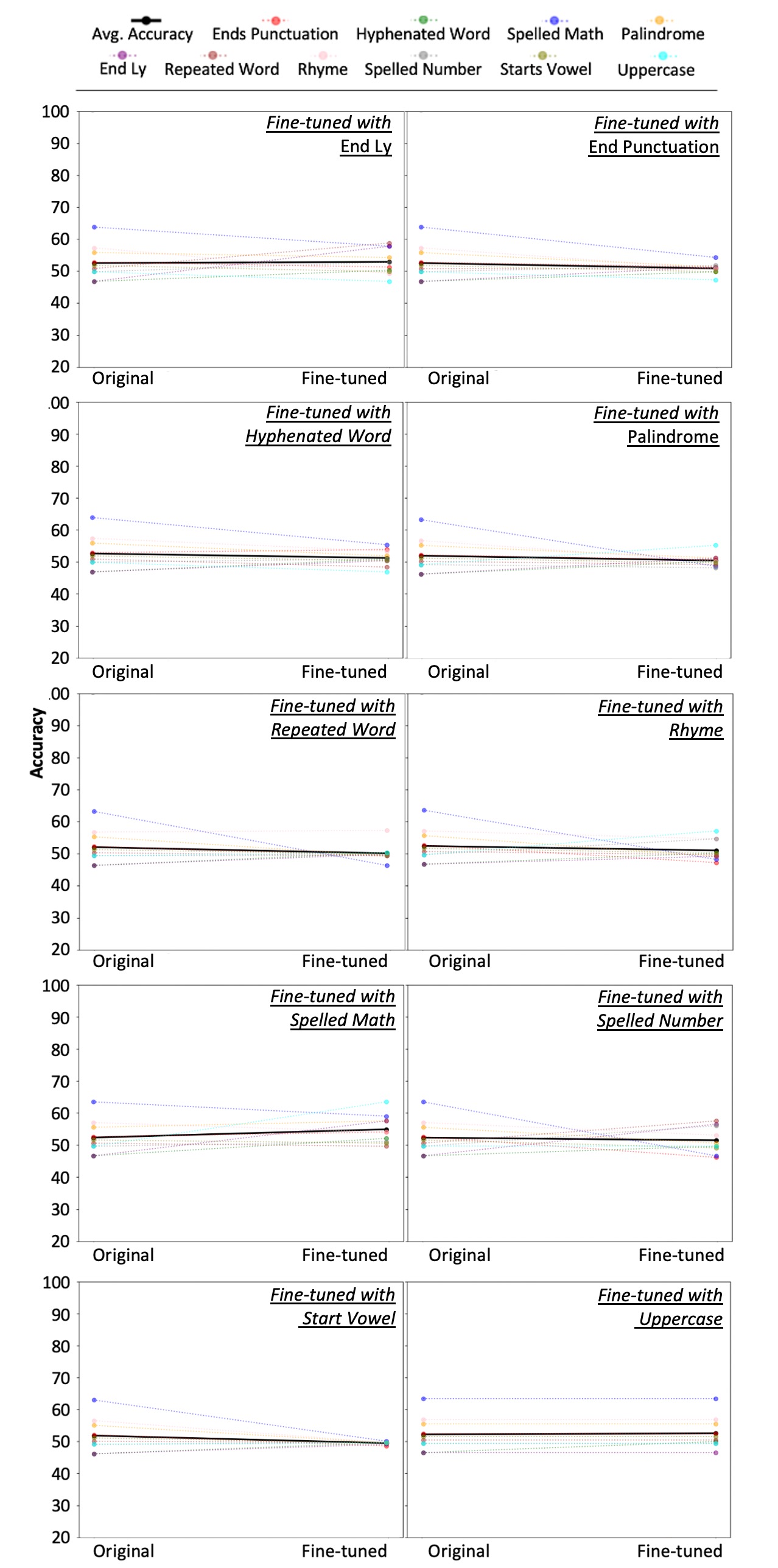}
\captionof{figure}{\textbf{GPT-3.5 Doesn't Solve H-Test Better with Fine-Tuning}: We report gpt-3.5-turbo-0613 original vs. fine-tuned performances across \textsc{H-Test}. Appendix \ref{App:FurtherTrain}.}
\label{fig:finetune}
\vspace{-4mm}
\end{figure}

\begin{table*}[ht]
\centering
\footnotesize
\begin{tabular}{l|ccc cccc ccc | c}
\textbf{Model} & \rotatebox{90}{\textbf{End Ly}} & \rotatebox{90}{\textbf{End Punctuation}} & \rotatebox{90}{\textbf{Hyphenated Word}} & \rotatebox{90}{\textbf{Palindrome}} & \rotatebox{90}{\textbf{Repeated Word}} & \rotatebox{90}{\textbf{Rhyme}} & \rotatebox{90}{\textbf{Spelled Math}} & \rotatebox{90}{\textbf{Spelled Number}} & \rotatebox{90}{\textbf{Start Vowel}} & \rotatebox{90}{\textbf{Uppercase}} & \textbf{Average}\\
\midrule
\midrule
\multicolumn{12}{l}{A. \textbf{Scaling} | Did Not Improve} \\
\midrule
\midrule
LLaMA 2 13B & 50.5 & 45.5 & 45.0 & 51.0 & 51.0 & 50.0 & 51.0 & 50.0 & 47.5 & 50.0 & \textbf{49.2}\\
LLaMA 2 70B & 45.5 & 49.0 & 47.5 & 40.0 & 46.0 & 45.5 & 45.0 & 47.0 & 32.5 & 53.5 & \textbf{44.8}\\
\midrule
\midrule
\multicolumn{12}{l}{B. \textbf{Multi-Modal Expansion} | Performance Improved But Likely Due to Stronger Instruction Following} \\
\midrule
\midrule
N.H. Yi 34B & 53.0 & 27.0 & 42.5 & 4.0 & 6.0 & 26.0 & 14.0 & 39.5 & 55.5 & 39.0 & \textbf{30.7}\\
LLaVA 34B     & 63.0 & 53.5 & 51.5 & 53.0 & 47.5 & 59.0 & 53.0 & 54.5 & 57.5 & 56.0 & \textbf{54.9}\\
\midrule
\midrule
\multicolumn{12}{l}{C. \textbf{Unknown Proprietary Method - 1} | Performance Improved and ``Solves''} \\
\midrule
\midrule
Claude V2 & 51.5 & 72.0 & 56.5 & 61.5 & 69.5 & 63.0 & 47.5 & 78.5 & 57.5 & 62.5 & \textbf{62.2}\\
Claude 3 Opus & 68.5 & 89.0 & 74.5 & 79.5 & 86.5 & 93.5 & 77.0 & 68.5 & 69.0 & 73.5 & \textbf{76.2}\\
\midrule
\midrule
\multicolumn{12}{l}{D. \textbf{Unknown Proprietary Method - 2} | Performance Improved and ``Solves''} \\
\midrule
\midrule
GPT 3.5 & 49.5 & 51.5 & 52.5 & 55.5 & 46.5 & 63.5 & 49.5 & 57.0 & 50.5 & 46.5 & \textbf{52.3}\\
GPT 4o & 84.5 & 75.0 & 72.5 & 94.0 & 73.0 & 94.5 & 73.0 & 56.0 & 68.5 & 59.0 & \textbf{75.0}\\
\bottomrule
\end{tabular}
\caption{\textbf{Different Approaches and Their Impact on \textsc{H-Test} Performance}: We report comparisons across different modeling approaches: scaling (LLaMA 2), multi-modal expansion (N.H. Yi 34B to LLaVA 34B), and proprietary methods (Claude and GPT). The table highlights that while some approaches lead to improvements, none consistently solve the \textsc{H-Test}.}
\vspace{-6mm}
\label{tab:multimodal}
\end{table*}

We do not understand GPT-4o and Claude 3 Opus's performance on \textsc{H-Test}. We had three hypotheses aimed at replicating this success: \textbf{1. Data:} training more orthography-specific data can improve \textsc{H-Test} score, \textbf{2. Modality:} vision-modality can improve \textsc{H-Test} score, \textbf{3. Architecture:} mixture-of-experts (MoE) architecture \citep{jiang2024mixtral} can improve \textsc{H-Test} score. However, we fail at pinpointing what really made solving \textsc{H-Test} possible.

\textbf{Analysis 1. Training with more orthography-specific language data does not improve \textsc{H-Test}.} This is our strongest observation so far that proves LLMs don't naturally learn to solve \textsc{H-Test} tasks. We produced 1000 training instances per task in \textsc{H-Test}, and fine-tuned gpt-3.5-turbo-0613 ten different times accordingly. After training for three epochs on each task, we evaluate them on \textsc{H-Test} at $k = 50$ and observe that no significant performance improvement was achieved (Figure \ref{fig:finetune}; Appendix \ref{App:FurtherTrain}). Across all tasks, the performance change (in both directions) was not statistically significant, suggesting that fine-tuning on orthographically rich data does not bridge the sensory experience gap.

\textbf{Analysis 2. Multi-modality does not automatically improve \textsc{H-Test} performance.} In Table \ref{tab:multimodal}, we present a performance comparison between Nous Hermes Yi 34B and LLaVA V1.6 34B \citep{liu2023visual}, the latter being a vision-language model based on the former. At the time of writing, LLaVA V1.6 34B is the strongest open-source multi-modal model available. Despite the addition of visual modality, we observe that simply incorporating visual data into the training does not result in a straightforward improvement in \textsc{H-Test} performance. LLaVA V1.6 showed around random-chance performance, which is clearly not an indication of the model's ability to solve \textsc{H-Test}.

We initially planned on running additional MoE experiments using instruction-tuned Mistral-7B-Instruct-V2 \citep{jiang2023mistral} and Mixtral-8X7B-Instruct-V1 \citep{jiang2024mixtral} but we skipped this evaluation as the models did not understand that task (at k = 50) more than 70\% of the test cases. More about this difficulty of evaluating \textsc{H-Test} on smaller models is discussed in Appendix \ref{App:responsesandopen}.

We could not specify a hypothesis that singlehandedly and meaningfully improved \textsc{H-Test} performance. Fortunately, these results strengthen our initial claim on the limitations of language-only models, but we fail to explain how GPT-4o and Claude 3 Opus are solving \textsc{H-Test}. 
    \section{Background}
    \textbf{Philosophical Case of Experiential Knowledge}
    The philosophical debate on the nature of knowledge, especially the contrast between experiential and propositional knowledge, is vividly illustrated in Jackson's ``Mary's Room'' thought experiment. Jackson (1986) contends that propositional knowledge is inadequate without experiential knowledge. This is exemplified by Mary, who, despite her extensive propositional knowledge about color, gains a new type of understanding only upon experiencing color firsthand \citep{jackson1986mary, jackson1982epiphenomenal}. This thought experiment has sparked extensive discourse in philosophy, particularly around the concept of qualia, which refers to individual instances of conscious experience \citep{dennett1993consciousness}.

    Thomas Nagel, in his paper ``What is it Like to be a Bat?'', explores the subjective nature of experience, arguing that an objective understanding of the mind must account for the subjective character of experience \citep{nagel1980like}. This work is pivotal in understanding the limits of propositional knowledge and the importance of experiential understanding. In a similar line of work, \citet{chalmers1997conscious} discusses the `hard problem' of consciousness, emphasizing the gap between explanatory physical processes and the qualitative experience of consciousness \citep{chalmers1997conscious}. This reinforces that propositional knowledge might not be sufficient in fully explaining conscious experience.

    \citet{dennett1993consciousness}, offers a contrasting view, arguing that subjective experiences (qualia) can be explained in terms of physical processes and brain functions \citep{dennett1993consciousness}. This perspective is crucial in the debate over whether experiential knowledge is any different from propositional knowledge.

    \textbf{Large Language Models and Language Tasks}
    The development and rise of large language models (LLMs) like GPT-3 have brought about significant advancements in the field of natural language processing. These models, however, process language fundamentally differently from humans, who experience language through sensory inputs \citep{bender2020climbing, brown2020language}. This difference in processing mechanisms poses challenges in comprehensively modeling human language understanding \citep{hinton2015distilling, russell2010artificial}. Text-only training of LLMs, while effective for many applications, shows limitations in capturing the full essence of human language understanding, especially in tasks requiring sensory and stylistic comprehension \citep{manning1999foundations}. Such issues has been investigated before but we believe our work is first to pinpoint and quantify the orthographic deficiencies of LLMs \citep{rust2022language, belinkov2018synthetic}. These limitations underscore the gap in experiential knowledge inherent in current LLMs, as also demonstrated in this research.

    The concept of sensory deprivation in LLMs roughly parallels the experiences of visually impaired individuals. Unlike visually impaired persons, who often develop heightened abilities in other senses, LLMs lack such compensatory mechanisms\citep{likova2018transfer, amedi2017task}. This highlights the unique challenge in AI of developing models capable of processing and understanding language akin to human sensory experiences \citep{mitchell2010composition}.
    \section{Conclusion}
 Our research revisits the philosophical case of Mary's Room, illustrating that LLMs, much like Mary, lack a comprehensive understanding of language due to their sensory-deprived training. We empirically show that scaling model size and data, while beneficial for some aspects of language understanding, does not address the deficiency in orthographic abilities. This finding suggests that there are some fundamental limitations in the current language-modeling paradigm that prevent a holistic understanding of language and that it is still an open research direction. 

\section{Limitations}
It is important to recognize several limitations inherent to our research methodology and scope:

1.  The \textsc{H-Test} emphasizes visual and auditory capabilities and does not consider other sensory inputs like olfactory, gustatory, and tactile perceptions that influence human language understanding. Our conclusions are thus confined to the explored sensory attributes, potentially overlooking the integral role of these unexamined senses in comprehensive language processing.

2. Our analysis significantly relies on proprietary data, particularly the performance metrics and training methodologies of language models like GPT-4. The proprietary nature of these models limits our ability to conduct independent verification, introducing a dependency on the accuracy and integrity of third-party reporting.

3. The assumption that multi-modal training inherently resolves the sensory gap is also a simplification. The integration of multi-modal data into language model training is a complex task that may introduce new challenges and biases, which our study does not address in detail.

4. Our application of philosophical concepts such as qualia to AI and language models operates metaphorically. The subjective nature of qualia and its implications for consciousness in humans do not translate directly to the objective, computational processes of language models and no anthropomorphic generalization should be made.

5. It is also crucial to note that the field of AI and language modeling is rapidly evolving. The capabilities of language models are continually advancing, and newer models may exhibit sensory processing abilities not captured by the current iteration of the \textsc{H-Test}. Our study establishes a dichotomy between human and machine processing of language, not accounting for the possibility that language models may develop distinct, non-human sensory processing methods that are effective in their unique computational context.

\newpage

\bibliography{custom}
\bibliographystyle{acl_natbib}

    \clearpage
    \appendix
    \onecolumn
\section{Full Obtained Results}
\label{App:fullResults}
\begin{table*}[h]
\centering
\footnotesize
\resizebox{\textwidth}{!}{%
\begin{tabular}{l|c|c|c|c |c|c|c|c|c |c|c|c|c|c |c}

\textbf{Task} 
& \rotatebox{90}{\textbf{J2 Mid}} 
& \rotatebox{90}{\textbf{J2 Ultra}} 
& \rotatebox{90}{\textbf{Titan Lite}} 
& \rotatebox{90}{\textbf{Claude Instant V1}} 

& \rotatebox{90}{\textbf{Claude V1}} 
& \rotatebox{90}{\textbf{Claude V2}} 
& \rotatebox{90}{\textbf{Command-Light}}
& \rotatebox{90}{\textbf{Command}} 
& \rotatebox{90}{\textbf{GPT 3.5}} 

& \rotatebox{90}{\textbf{GPT 4o}} 
& \rotatebox{90}{\textbf{LLaMA 2 13B}} 
& \rotatebox{90}{\textbf{LLaMA 2 70B}} 
& \rotatebox{90}{\textbf{Luminous Base}} 
& \rotatebox{90}{\textbf{Luminous Extended}} 

& \rotatebox{90}{\textbf{Luminous Supreme}} \\ 

\midrule
\midrule
\multicolumn{16}{c}{\textbf{k = 50, \textsc{H-Test}}} \\
\midrule

Uppercase       & 0.46   & 0.505     & 0.5         & 0.535           & 0.61      & 0.555     & 0.455         & 1.0      & 0.495    & - & 0.505     & 0.455       & 0.51        & 0.51          & 0.49                  \\ 
Start Vowel     & 0.47   & 0.51      & 0.5         & 0.58            & 0.665     & 0.65      & 0.475         & 0.5      & 0.515    & -  & 0.455     & 0.49        & 0.495       & 0.525         & 0.49                  \\ 
Ends Punctuation& 0.505  & 0.48      & 0.52        & 0.705           & 0.67      & 0.52      & 0.545         & 1.0      & 0.525    & - & 0.45      & 0.475       & 0.49        & 0.515         & 0.52                  \\ 
Palindrome      & 0.51   & 0.51      & 0.51        & 0.54            & 0.6       & 0.58      & 0.205         & 0.5      & 0.555    & -  & 0.51      & 0.4         & 0.485       & 0.505         & 0.505                 \\ 
End Ly          & 0.61   & 0.585     & 0.5         & 0.58            & 0.75      & 0.75      & 0.61          & 1.0      & 0.465    & -  & 0.51      & 0.46        & 0.5         & 0.525         & 0.51                  \\ 
Spelled Math    & 0.495  & 0.51      & 0.5         & 0.63            & 0.6       & 0.695     & 0.405         & 0.5      & 0.635    & - & 0.5       & 0.455       & 0.51        & 0.515         & 0.505                 \\ 
Spelled Number  & 0.515  & 0.51      & 0.5         & 0.58            & 0.555     & 0.55      & 0.49          & 0.5      & 0.495    & - & 0.51      & 0.45        & 0.505       & 0.51          & 0.495                 \\ 
Rhyme           & 0.51   & 0.495     & 0.49        & 0.685           & 0.895     & 0.775     & 0.51          & 0.5      & 0.57     & - & 0.5       & 0.47        & 0.515       & 0.51          & 0.515                 \\ 
Repeated Word   & 0.515  & 0.495     & 0.485       & 0.625           & 0.69      & 0.53      & 0.54          & 0.5      & 0.505    & -  & 0.475     & 0.325       & 0.51        & 0.515         & 0.51                  \\ 
Hyphenated Word & 0.405  & 0.510     & 0.5         & 0.58            & 0.615     & 0.58      & 0.47          & 0.5      & 0.465    & -  & 0.5       & 0.535       & 0.52        & 0.515         & 0.5                     \\ 
\hline
Average         & 0.5    & 0.511     & 0.501       & 0.604           & 0.665     & 0.619     & 0.471         & 0.65     & 0.523    & - & 0.4915    & 0.448       & 0.5025      & 0.5125        & 0.504                 \\
\midrule
\midrule
\multicolumn{16}{c}{\textbf{k = 28, \textsc{H-Test}}} \\
\midrule
Uppercase       & -       & 0.51      & -          & -              & 0.58      & 0.48      & -             & 0.51    & 0.53     & -     & -          & 0.49        & -            & -               & -                       \\ 
Start Vowel     & -       & 0.525     & -          & -              & 0.66      & 0.61      & -             & 0.505   & 0.51     & -     & -          & 0.485      & -            & -               & -                       \\ 
Ends Punctuation& -       & 0.515     & -          & -              & 0.74      & 0.585     & -             & 0.645   & 0.505    & -     & -          & 0.44       & -            & -               & -                       \\ 
Palindrome      & -       & 0.505     & -          & -              & 0.675     & 0.595     & -             & 0.49    & 0.52     & -     & -          & 0.215      & -            & -               & -                       \\ 
End Ly          & -       & 0.525     & -          & -              & 0.76      & 0.715     & -             & 0.625   & 0.545    & -     & -          & 0.45       & -            & -               & -                       \\ 
Spelled Math    & -       & 0.515     & -          & -              & 0.595     & 0.665     & -             & 0.5     & 0.635    & -     & -          & 0.435      & -            & -               & -                       \\ 
Spelled Number  & -       & 0.51      & -          & -              & 0.53      & 0.525     & -             & 0.495   & 0.59     & -     & -          & 0.42       & -            & -               & -                       \\ 
Rhyme           & -       & 0.51      & -          & -              & 0.78      & 0.845     & -             & 0.555   & 0.44     & -     & -          & 0.395      & -            & -               & -                       \\ 
Repeated Word   & -       & 0.515     & -          & -              & 0.66      & 0.525     & -             & 0.58    & 0.52     & -     & -          & 0.285      & -            & -               & -                       \\ 
Hyphenated Word & -       & 0.515     & -          & -              & 0.58      & 0.585     & -             & 0.585   & 0.555    & -     & -          & 0.485      & -            & -               & -                       \\ 
\hline
Average         & -       & 0.5141    & -          & -              & 0.6436    & 0.5994    & -             & 0.543   & 0.5314   & -     & -          & 0.4014     & -            & -               & -                       \\

\midrule
\midrule
\multicolumn{16}{c}{\textbf{k = 14, \textsc{H-Test}}} \\
\midrule
Uppercase       & -       & 0.5       & -          & -              & 0.58      & 0.515     & -             & 0.5     & 0.52     & 0.845 & -          & 0.475       & -            & -               & -                       \\ 
Start Vowel     & -       & 0.515     & -          & -              & 0.685     & 0.72      & -             & 0.505   & 0.46     & 0.75 & -          & 0.495      & -            & -               & -                       \\ 
Ends Punctuation& -       & 0.525     & -          & -              & 0.7       & 0.565     & -             & 0.67    & 0.48     & 0.725  & -          & 0.435      & -            & -               & -                       \\ 
Palindrome      & -       & 0.51      & -          & -              & 0.62      & 0.615     & -             & 0.505   & 0.525    & 0.94 & -          & 0.21       & -            & -               & -                       \\ 
End Ly          & -       & 0.595     & -          & -              & 0.735     & 0.695     & -             & 0.615   & 0.495    & 0.73 & -          & 0.445      & -            & -               & -                       \\ 
Spelled Math    & -       & 0.555     & -          & -              & 0.56      & 0.63      & -             & 0.495   & 0.62     & 0.945 & -          & 0.435      & -            & -               & -                       \\ 
Spelled Number  & -       & 0.5       & -          & -              & 0.58      & 0.475     & -             & 0.515   & 0.53     & 0.73 & -          & 0.445      & -            & -               & -                       \\ 
Rhyme           & -       & 0.505     & -          & -              & 0.78      & 0.785     & -             & 0.58    & 0.475    & 0.56  & -          & 0.43       & -            & -               & -                       \\ 
Repeated Word   & -       & 0.505     & -          & -              & 0.68      & 0.575     & -             & 0.565   & 0.46     & 0.685  & -          & 0.32       & -            & -               & -                       \\ 
Hyphenated Word & -       & 0.51      & -          & -              & 0.575     & 0.625     & -             & 0.555   & 0.51     & 0.59 & -          & 0.485      & -            & -               & -                       \\ 
\hline
Average         & -       & 0.516     & -          & -              & 0.6561    & 0.622     & -             & 0.556   & 0.511    & 0.75 & -          & 0.3987     & -            & -               & -                       \\

\midrule
\midrule
\multicolumn{16}{c}{\textbf{k = 4, \textsc{H-Test}}} \\
\midrule
Uppercase       & -       & 0.5       & -          & -              & 0.54      & 0.525     & -             & 0.495   & 0.565    & -     & -          & 0.48        & -            & -               & -                       \\ 
Start Vowel     & -       & 0.52      & -          & -              & 0.65      & 0.61      & -             & 0.51    & 0.48     & -     & -          & 0.5         & -            & -               & -                       \\ 
Ends Punctuation& -       & 0.5       & -          & -              & 0.695     & 0.545     & -             & 0.68    & 0.53     & -     & -          & 0.42       & -            & -               & -                       \\ 
Palindrome      & -       & 0.505     & -          & -              & 0.605     & 0.62      & -             & 0.485   & 0.49     & -     & -          & 0.235      & -            & -               & -                       \\ 
End Ly          & -       & 0.57      & -          & -              & 0.695     & 0.725     & -             & 0.59    & 0.46     & -     & -          & 0.445      & -            & -               & -                       \\ 
Spelled Math    & -       & 0.505     & -          & -              & 0.57      & 0.68      & -             & 0.51    & 0.55     & -     & -          & 0.44       & -            & -               & -                       \\ 
Spelled Number  & -       & 0.505     & -          & -              & 0.62      & 0.5       & -             & 0.49    & 0.555    & -     & -          & 0.43       & -            & -               & -                       \\ 
Rhyme           & -       & 0.5       & -          & -              & 0.775     & 0.845     & -             & 0.565   & 0.375    & -     & -          & 0.42       & -            & -               & -                       \\ 
Repeated Word   & -       & 0.5       & -          & -              & 0.705     & 0.52      & -             & 0.535   & 0.49     & -     & -          & 0.29       & -            & -               & -                       \\ 
Hyphenated Word & -       & 0.5       & -          & -              & 0.605     & 0.605     & -             & 0.59    & 0.515    & -     & -          & 0.48       & -            & -               & -                       \\ 
\hline
Average         & -       & 0.5035    & -          & -              & 0.6495    & 0.6235    & -             & 0.55    & 0.503    & -     & -          & 0.3995     & -            & -               & -                       \\

\midrule
\midrule
\midrule
\multicolumn{16}{c}{\textbf{Letter Geometry}} \\
\midrule
Letter Geometry  & -    & -    & -    & 0.378 & 0.400 & 0.289 & 0.134 & 0.144 & 0.444 & -  & -    & -    & -    & -   & -    \\ 

\bottomrule
\end{tabular}}
\caption{\textbf{Full Results}: We report full obtained results across k = \{50, 28, 14, 4\} on \textsc{H-Test} and Letter Geometry. *Claude 3 Opus results in Table \ref{tab:multimodal}.}
\label{tab:full}
\vspace{-4mm}
\end{table*}

\section{Experimental Details}
\label{App:abDetail}
We accessed all APIs ("anthropic.claude-instant-v1", "anthropic.claude-v1", "anthropic.claude-v2", "ai21.j2-mid-v1", "ai21.j2-ultra-v1", "command-light", "command", "gpt-3.5-turbo-0613", "gpt-4-0613", "luminous-base", "luminous-exteneded", "luminous-supreme", "amazon.titan-text-lite-v1", "llama 2 13B", "llama 2 70B") between the third week of November and the third week of December 2023. We accessed "gpt-4o-2024-05-13" and "claude-3-opus-20240229" in the first week of June 2024. In particular, we access LLaMA through Replicate, and J2, Claude, and Titan through Amazon Bedrock. All other APIs were accessed through the respective providers. For all experiments, the model temperature was set at 0.7, and the random seed (for generating data) was set at 12062023. The results and exact dataset used in this research are recoverable by setting the random seed to 12062023. All few-shot tests are conducted with max\_new\_tokens parameters set to five. Full code and data at <github.com/brucewlee/h-test>. 

\section{On Model Responses and Small Open-Source Models}
\label{App:responsesandopen}
In our experiments, we observed that major models, including ChatGPT, produced responses in the intended format of a single letter (A or B), as explicitly instructed in our prompts. Indeed, this adherence to the output format was a critical aspect of our experimental design to ensure consistency in the evaluation of model responses. As documented in Appendix C of our paper, each prompt concluded with the directive ``(Respond in one letter and nothing else)'', which was intended to guide the models toward the desired response format.

Smaller, open-source models (tested: LLaMA 7B, LLaVA, Mistral, Mixtral, etc.) almost always didn't understand the task (didn't respond in A or B). This lack in in-context learning ability is reminiscent of a recent work \citep{lu2023emergent}. They showed a tendency to generate responses that were either irrelevant or completely off-topic, which we categorized as ``gibberish.'' Due to this significant deviation from the expected task performance, these models were subsequently excluded from our final analysis. This decision was based on the premise that their responses did not provide meaningful data for evaluating the specific language understanding capabilities we were investigating, as the response of the language model that could not process the requirements of the task was irrelevant to our investigation.

The issue of difficult-to-parse response was pronounced in our Chain-of-Thought (CoT) experiment setup. In this setup, models were prompted to generate a chain of thought before arriving at a conclusion (A or B). We found that this approach sometimes led to responses that did not reach a specific answer. To account for this discrepancy, we reported two different accuracy metrics in Figure 4 of our paper: ``Average Accuracy'' and ``Adjusted Accuracy''. The ``Adjusted Accuracy'' metric specifically addressed cases where models provided the correct answer but not in the prescribed single-letter format.

\section{More on Parsing CoT Model Responses}
\label{App:CoT}
In implementing this CoT setup, we follow the conventional CoT evaluation setup where an LLM to be tested generates a lengthy CoT response (which is contrastive to the one-letter response in the previous setups) and another LLM reads and determines which option (A or B) the lengthy response is pointing to \citep{chen2023you}.
This two-step process inevitably generates cases where the latter LLM cannot clearly interpret what the former LLM's response pointed to, and we call these cases out-of-context responses. 
The gray line in Figure \ref{fig:cotcomparison} shows adjusted accuracy, excluding these out-of-context responses. 
Hence, the CoT@14 datapoint's adjusted accuracy (in gray) represents an accuracy of less than 200 test instances per subtask.

\section{Human Baseline Performance on \textsc{H-Test}}
\label{App:HumanBaseline}
\begin{table}[h]
\centering
\footnotesize
\resizebox{0.4\textwidth}{!}{%
\begin{tabular}{l|c|c|c|c|c|c}

\textbf{Task} & \textbf{k = 4} & \textbf{k = 14} & \textbf{k = 28} & \textbf{k = 50}\\ 

\midrule

Human A        & 10/10  & -   & -   & -   \\ 
Human B        & -    & 10/10   & -   & -   \\ 
Human C        & -    & -   & 10/10   & -   \\ 
Human D        & -    & -   & -   & 10/10   \\ 

\bottomrule
\end{tabular}}
\caption{\textbf{Human Performance:} We report human performance in identifying the patterns behind the given few-shot examples in A/B classification tasks in \textsc{H-Test}. The numbers indicate how many correct rules were identified out of ten \textsc{H-Test} tasks. } 
\label{tab:human}
\vspace{-4mm}
\end{table}

In light of the \textsc{H-Test}'s design, which is readily and trivially solvable once the classification pattern behind the given few-shot examples is understood, our approach to establishing a human baseline focused not on direct label classification as was done with the language models but on the participants' ability to identify and verbally articulate these patterns.

\textbf{Participants:} Four English-native undergraduate students from a US university were recruited. \textbf{Procedure:} Participants were individually presented with the same few-shot examples given to LLMs from the H-Test, printed on paper. Participants were asked to identify the underlying pattern or rule that distinguished Group A from Group B examples. Upon identifying the pattern, participants were instructed to verbally articulate this rule as precisely as possible. \textbf{Data Collection:} We only considered a participant's answer right when it directly matched the actual classification criteria designed for each task. We consider a response wrong if the participant took more than three minutes. We told the participants that ``sentences A and B are different based on a linguistic rule''. If the participant asked for clarification, we responded ``linguistic rules are patterns like ... (Starts with a certain pattern ...)''. No further clarification was provided.

\section{Fine-tuned Performance, GPT-3.5 on \textsc{H-Test}}
\label{App:FurtherTrain}

\begin{table}[ht]
\centering
\footnotesize
\resizebox{0.75\textwidth}{!}{%
\begin{tabular}{l|l|llllllllll}

\rotatebox{45}{\textbf{\hspace{3mm}gpt-3.5-turbo-0613}} & \rotatebox{90}{\textbf{Original}} & \rotatebox{90}{\textbf{Uppercase}} & \rotatebox{90}{\textbf{Start Vowel}} & \rotatebox{90}{\textbf{End Punctuation}} & \rotatebox{90}{\textbf{Palindrome}} & \rotatebox{90}{\textbf{End Ly}} & \rotatebox{90}{\textbf{Spelled Math}} & \rotatebox{90}{\textbf{Spelled Number}} & \rotatebox{90}{\textbf{Rhyme}} & \rotatebox{90}{\textbf{Repeated Word}} & \rotatebox{90}{\textbf{Hyphenated Word}} \\ 
\midrule
Uppercase        & 49.5 & \underline{\textbf{50.}}  & 47.   & 50.5 & 51.  & 49.  & 47.5 & 56.6 & 51.  & 46.  & 50. \\
Start Vowel      & 51.5 & 50.  & \underline{\textbf{51.5}}  & 49.5 & 49.5 & 49.5 & 48.5 & 49.  & 50.  & 49.  & 50. \\
End Punctuation  & 52.5 & 50.  & 49.   & \underline{\textbf{51.}}  & 50.  & 51.  & 50.  & 50.  & 46.  & 49.5 & 49.5 \\
Palindrome       & 55.5 & 50.  & 51.   & 51.  & \underline{\textbf{51.5}} & 51.5 & 49.  & 48.5 & 50.5 & 49.5 & 50.5 \\
End Ly           & 46.5 & 49.5 & 49.   & 51.  & 54.  & \underline{\textbf{57.5}} & 50.  & 52.5 & 56.5 & 50.  & 50. \\
Spelled Math     & 63.5 & 50.5 & 32.5  & 54.  & 57.5 & 59.  & \underline{\textbf{53.}}  & 51.  & 54.  & 46.5 & 52. \\
Spelled Number   & 49.5 & 50.  & 46.   & 51.5 & 53.5 & 46.5 & 48.5 & \underline{\textbf{56.}}  & 53.  & 49.5 & 50.5 \\
Rhyme            & 57.  & 50.  & 47.   & 50.5 & 49.  & 49.  & 48.  & 54.5 & \underline{\textbf{57.}}  & 49.5 & 50. \\
Repeated Word    & 50.5 & 50.  & 50.   & 50.5 & 58.5 & 49.5 & 45.  & 52.  & 57.5 & \underline{\textbf{49.5}} & 50. \\
Hyphenated Word  & 46.5 & 50.  & 53.5  & 51.5 & 49.5 & 54.  & 55.  & 49.  & 53.  & 48.  & \underline{\textbf{50.5}} \\ 
\midrule
Average          & 52.3 & 50.  & 47.65 & 51.1 & 52.4 & 51.7 & 49.5 & 51.9 & 52.9 & 48.7 & 50.3 \\ 
\bottomrule
\end{tabular}}
\caption{\textbf{\textsc{H-Test} cannot be conveniently solved with more data}: Performance of gpt-3.5-turbo-0613 before and after further training with various \textsc{H-Test} tasks ($\sim$ 1000 train instances each task). All tests were conducted under $k = 50$ setup. Underscored values represent the cases where the training data task (column) was the same as the test data task (row), which is often conveniently referred to as in-domain.}
\label{tab:performance_enhancements}
\end{table}

In our efforts to understand the limits of language-only models in processing tasks that require sensory experience, we conducted a series of fine-tuning experiments on GPT-3.5, specifically targeting its performance on the \textsc{H-Test}. We selected GPT-3.5 for fine-tuning due to its advanced capabilities and widespread use in both academic and industrial settings. The model was fine-tuned using a custom dataset derived from the H-TEST, comprising 1,000 instances for each task. Fine-tuning was performed over three epochs on the official OpenAI platform. Across all tasks, the improvements were not significant, suggesting that fine-tuning on orthographically rich data does not bridge the sensory experience gap.

\newpage

\section{Task Exemplars}
\label{App:abExemplar}
\noindent\texttt{\textbf{\textsc{H-Test}: Ends Punctuation}}
\begin{tcolorbox}[breakable, enhanced]

Input: "Engine on the wall Octopus..." Label: A

Input: "Ice cream ! in the forest Teacher" Label: B

Input: "Island in the morning Elevator!" Label: A

Input: "Owl with interest ... Friend" Label: B

Input: "Penguin with precision River?" Label: A

Input: "Iguana all night ... Scientist" Label: B

Input: "Penguin to music Owl?" Label: A

Input: "Lion with interest ... Penguin" Label: B

Input: "Teacher with interest Dog!" Label: A

Input: "Cat ! to music Butterfly" Label: B

Input: "Cat in the hall Insect..." Label: A

Input: "Astronaut on the wall ... Opera" Label: B

Input: "Owl in the morning Book!" Label: A

Input: "Insect ! in the studio Book" Label: B

Input: "Lion in the kitchen House!" Label: A

Input: "Insect in the hall ? Flower" Label: B

Input: "Friend very quickly House!" Label: A

Input: "Butterfly to music ... Penguin" Label: B

Input: "Penguin in the forest Astronaut?" Label: A

Input: "Apple ? every second Car" Label: B

Input: "School in the sky Neighbor..." Label: A

Input: "Orchestra on the wall ... School" Label: B

Input: "Dog with passion River?" Label: A

Input: "Octopus ... in the forest Urchin" Label: B

Input: "Elephant in the park Urchin?" Label: A

Input: "Teacher to music ... Car" Label: B

Input: "Mountain in the forest Car!" Label: A

Input: "Albatross in the forest . Butterfly" Label: B

Input: "Teacher in the morning Butterfly." Label: A

Input: "Octopus in the morning ! Flower" Label: B

Input: "Albatross with precision Butterfly..." Label: A

Input: "Penguin in the studio ! Giraffe" Label: B

Input: "House in space Butterfly?" Label: A

Input: "Dog ? on the wall River" Label: B

Input: "Orchestra in the park Neighbor!" Label: A

Input: "Teacher in space ... Penguin" Label: B

Input: "Unicorn through the telescope Insect..." Label: A

Input: "Dog ? in the park Engine" Label: B

Input: "Owl to music Owl..." Label: A

Input: "Mountain in the sky ? Teacher" Label: B

Input: "Neighbor very quickly Urchin." Label: A

Input: "Umbrella ? along the valley Car" Label: B

\textbf{Input: "Kangaroo in the park Umbrella." Label: }

\textbf{A }

\textbf{B (Respond in one letter and nothing else)}
\end{tcolorbox}
\newpage

\noindent\texttt{\textbf{\textsc{H-Test}: Uppercase}}
\begin{tcolorbox}[breakable, enhanced]

Input: "engine jumps all night aCcurately." Label: A

Input: "penguin runs all night intently." Label: B

Input: "rivEr drives with precision eagerly." Label: A

Input: "tree sleeps in space quietly." Label: B

Input: "owl obserVes delicious meals slowly." Label: A

Input: "Penguin flows in the studio happily." Label: B

Input: "apple Listens in the studio expertly." Label: A

Input: "airport calculates in the studio enthusiastically." Label: B

Input: "unicorn obServes very quickly fast." Label: A

Input: "owl reads along the valley skillfully." Label: B

Input: "airport travels very quicklY melodiously." Label: A

Input: "Insect dances along the valley carefully." Label: B

Input: "insecT paints in the hall attentively." Label: A

Input: "tree sleeps all night adventurously." Label: B

Input: "friend runs in the morning sKillfully." Label: A

Input: "book paints in the studio attentively." Label: B

Input: "opera sings in the park eAgerly." Label: A

Input: "Insect paints in the kitchen creatively." Label: B

Input: "opera groWs in the forest brightly." Label: A

Input: "owl grows all night gracefully." Label: B

Input: "ice cream groWs in the forest carefully." Label: A

Input: "Umbrella talks to music attentively." Label: B

Input: "urchin daNces in the sky adventurously." Label: A

Input: "school ticks to the city attentively." Label: B

Input: "school reaDs with interest happily." Label: A

Input: "kangaroo glows through the telescope attentively." Label: B

Input: "umbrella teaches in the park hapPily." Label: A

Input: "opera grows in space consistently." Label: B

Input: "Urchin glows delicious meals steadily." Label: A

Input: "book runs in the forest gracefully." Label: B

Input: "girAffe jumps with interest carefully." Label: A

Input: "engine jumps in the hall expertly." Label: B

Input: "elevator cooks on the wall eagerlY." Label: A

Input: "friend ticks along the valley skillfully." Label: B

Input: "albatross Calculates in the forest gracefully." Label: A

Input: "cat observes delicious meals melodiously." Label: B

Input: "orchestra jumps tO music consistently." Label: A

Input: "octopus grows with interest smoothly." Label: B

Input: "book jumps In the morning skillfully." Label: A

Input: "umbrella paints in the sky smoothly." Label: B

Input: "butterfly observes along the valley Skillfully." Label: A

Input: "butterfly listens in the forest gracefully." Label: B

Input: "Island flows with interest quietly." Label: A

Input: "opera cooks in space accurately." Label: B

\textbf{Input: "tree paints on the wall accurately." Label: }

\textbf{A }

\textbf{B (Respond in one letter and nothing else)}
\end{tcolorbox}
\newpage

\noindent\texttt{\textbf{\textsc{H-Test}: Starts Vowel}}
\begin{tcolorbox}[breakable, enhanced]

Input: "Island sings to the city brightly." Label: A

Input: "Giraffe travels with passion melodiously." Label: B

Input: "Orchestra travels in the morning enthusiastically." Label: A

Input: "Robot paints in the park brightly." Label: B

Input: "Internet flows in the forest accurately." Label: A

Input: "Cupcake cooks through the telescope happily." Label: B

Input: "Olive drives with precision eagerly." Label: A

Input: "Basket flows through the telescope adventurously." Label: B

Input: "Acrobat observes in the forest attentively." Label: A

Input: "Yacht paints in the park attentively." Label: B

Input: "Omelette runs in the morning skillfully." Label: A

Input: "River paints in the forest happily." Label: B

Input: "Airport ticks in the kitchen brightly." Label: A

Input: "Kite reads in the forest skillfully." Label: B

Input: "Octopus flows very quickly brightly." Label: A

Input: "Violin sleeps in the park steadily." Label: B

Input: "Elevator paints with passion accurately." Label: A

Input: "Car jumps in the morning steadily." Label: B

Input: "Artist travels along the valley fast." Label: A

Input: "Kangaroo cooks in space attentively." Label: B

Input: "Umbrella teaches in the park happily." Label: A

Input: "Bicycle cooks in the studio happily." Label: B

Input: "Eagle runs with interest gracefully." Label: A

Input: "Balloon sleeps in the hall smoothly." Label: B

Input: "Urchin glows delicious meals steadily." Label: A

Input: "Book jumps with passion eagerly." Label: B

Input: "Umpire jumps with interest carefully." Label: A

Input: "Car listens to music gracefully." Label: B

Input: "Acrobat sleeps with precision creatively." Label: A

Input: "Globe calculates on the wall melodiously." Label: B

Input: "Elevator cooks on the wall eagerly." Label: A

Input: "Computer observes to music expertly." Label: B

Input: "Economist grows every second happily." Label: A

Input: "Kite flows in the kitchen gracefully." Label: B

Input: "Albatross calculates in the forest gracefully." Label: A

Input: "Kangaroo listens in the hall melodiously." Label: B

Input: "Underdog dances with precision enthusiastically." Label: A

Input: "Mountain cooks in the morning quietly." Label: B

Input: "Orchestra jumps to music consistently." Label: A

Input: "Lion teaches with passion gracefully." Label: B

Input: "Explorer cooks in the sky adventurously." Label: A

Input: "Scientist listens on the wall quietly." Label: B

Input: "Umpire glows every second steadily." Label: A

Input: "Kangaroo grows in the hall intently." Label: B

\textbf{Input: "Lion calculates in the morning accurately." Label: }

\textbf{A }

\textbf{B (Respond in one letter and nothing else)}
\end{tcolorbox}
\newpage

\noindent\texttt{\textbf{\textsc{H-Test}: Spelled Number}}
\begin{tcolorbox}[breakable, enhanced]

Input: "Neighbor sings seven birds melodiously." Label: A

Input: "Flower reads 10 computers expertly." Label: B

Input: "Book grows sixteen birds fast." Label: A

Input: "Car drives 5 paintings adventurously." Label: B

Input: "Neighbor jumps thirteen books steadily." Label: A

Input: "Penguin grows 4 birds quietly." Label: B

Input: "Dog dances nine computers happily." Label: A

Input: "Butterfly ticks 9 computers enthusiastically." Label: B

Input: "Penguin observes seven songs brightly." Label: A

Input: "Giraffe paints 17 paintings melodiously." Label: B

Input: "Scientist cooks six computers brightly." Label: A

Input: "Mountain teaches 8 cars carefully." Label: B

Input: "Albatross paints five birds gracefully." Label: A

Input: "Artist sings 18 computers attentively." Label: B

Input: "Orchestra reads one books adventurously." Label: A

Input: "The car drives 19 apples brightly." Label: B

Input: "Book drives sixteen apples skillfully." Label: A

Input: "The cat calculates 12 computers enthusiastically." Label: B

Input: "Dog jumps five apples happily." Label: A

Input: "Lion dances 19 paintings intently." Label: B

Input: "House listens ten songs fast." Label: A

Input: "Insect cooks 18 trees eagerly." Label: B

Input: "The teacher listens twenty paintings intently." Label: A

Input: "Mountain observes 16 books adventurously." Label: B

Input: "The moon dances thirteen apples melodiously." Label: A

Input: "Penguin glows 16 cars gracefully." Label: B

Input: "Unicorn cooks seven apples adventurously." Label: A

Input: "Mountain cooks 14 paintings brightly." Label: B

Input: "Lion drives three apples steadily." Label: A

Input: "Opera calculates 5 songs adventurously." Label: B

Input: "The scientist reads two paintings smoothly." Label: A

Input: "The car sings 20 computers consistently." Label: B

Input: "House cooks seventeen birds attentively." Label: A

Input: "Astronaut observes 1 birds steadily." Label: B

Input: "Book jumps thirteen computers happily." Label: A

Input: "Orchestra reads 10 computers eagerly." Label: B

Input: "Insect observes seventeen apples fast." Label: A

Input: "The scientist reads 9 trees brightly." Label: B

Input: "Kangaroo paints sixteen birds smoothly." Label: A

Input: "Astronaut drives 11 paintings happily." Label: B

Input: "A dog runs five apples expertly." Label: A

Input: "Flower cooks 11 apples carefully." Label: B

Input: "Penguin reads nine books attentively." Label: A

Input: "Insect travels 17 paintings melodiously." Label: B

\textbf{Input: "Butterfly teaches 10 paintings fast." Label: }

\textbf{A }

\textbf{B (Respond in one letter and nothing else)}
\end{tcolorbox}
\newpage

\noindent\texttt{\textbf{\textsc{H-Test}: Rhyme}}
\begin{tcolorbox}[breakable, enhanced]

Input: "life pfeiff" Label: A

Input: "make meals" Label: B

Input: "way a." Label: A

Input: "get want" Label: B

Input: "think fincke" Label: A

Input: "hand feel" Label: B

Input: "work clerc" Label: A

Input: "passion place" Label: B

Input: "passion ashen" Label: A

Input: "park want" Label: B

Input: "work bourke" Label: A

Input: "person case" Label: B

Input: "night byte" Label: A

Input: "week meals" Label: B

Input: "person Christoffersen" Label: A

Input: "call hand" Label: B

Input: "year amir" Label: A

Input: "tell ask" Label: B

Input: "world hurled" Label: A

Input: "point leave" Label: B

Input: "park arc" Label: A

Input: "call have" Label: B

Input: "think inc" Label: A

Input: "year use" Label: B

Input: "call aul" Label: A

Input: "day group" Label: B

Input: "seem agleam" Label: A

Input: "place want" Label: B

Input: "make Jacque" Label: A

Input: "government do" Label: B

Input: "make haik" Label: A

Input: "night telescope" Label: B

Input: "do accrue" Label: A

Input: "give world" Label: B

Input: "go aglow" Label: A

Input: "government place" Label: B

Input: "eye akai" Label: A

Input: "person take" Label: B

Input: "park clarke" Label: A

Input: "park city" Label: B

Input: "think finck" Label: A

Input: "world quickly" Label: B

Input: "come ancrum" Label: A

Input: "find fact" Label: B

\textbf{Input: "give liv" Label: }

\textbf{A }

\textbf{B (Respond in one letter and nothing else)}
\end{tcolorbox}
\newpage

\noindent\texttt{\textbf{\textsc{H-Test}: Repeated Word}}
\begin{tcolorbox}[breakable, enhanced]

Input: "carefully quickly quickly Neighbor runs." Label: A

Input: "music Neighbor fast runs." Label: B

Input: "quietly teaches Scientist night. night." Label: A

Input: "Teacher enthusiastically jumps wall." Label: B

Input: "Fast Car Car wall runs." Label: A

Input: "enthusiastically wall Car observes." Label: B

Input: "music sleeps Dog Dog expertly." Label: A

Input: "jumps enthusiastically Scientist telescope." Label: B

Input: "quietly quietly She city talks." Label: A

Input: "listens He telescope fast." Label: B

Input: "Teacher happily meals meals observes." Label: A

Input: "drives intently He park." Label: B

Input: "night night Teacher reads fast." Label: A

Input: "runs The cat night slowly." Label: B

Input: "telescope Neighbor teaches teaches enthusiastically." Label: A

Input: "intently night talks Teacher." Label: B

Input: "teaches The cat cat meals attentively." Label: A

Input: "runs expertly He passion." Label: B

Input: "telescope eagerly eagerly She sleeps." Label: A

Input: "reads meals Book slowly." Label: B

Input: "meals meals jumps enthusiastically Book." Label: A

Input: "carefully runs Dog music." Label: B

Input: "city city reads quietly Book." Label: A

Input: "attentively drives He meals." Label: B

Input: "night sleeps Teacher enthusiastically. enthusiastically." Label: A

Input: "drives interest happily Book." Label: B

Input: "runs Car quietly quietly telescope." Label: A

Input: "quickly jumps enthusiastically Dog." Label: B

Input: "happily happily talks night Dog." Label: A

Input: "telescope observes She happily." Label: B

Input: "fast listens listens interest Book." Label: A

Input: "wall jumps He expertly." Label: B

Input: "fast wall wall Dog teaches." Label: A

Input: "fast The cat quickly sleeps." Label: B

Input: "enthusiastically enthusiastically Neighbor talks city." Label: A

Input: "slowly The cat cooks wall." Label: B

Input: "telescope fast Car Car talks." Label: A

Input: "music drives fast Neighbor." Label: B

Input: "Book wall fast fast jumps." Label: A

Input: "passion The cat slowly runs." Label: B

Input: "The cat cat jumps music fast." Label: A

Input: "interest The cat quietly talks." Label: B

Input: "observes observes wall attentively He." Label: A

Input: "drives interest Book enthusiastically." Label: B

\textbf{Input: "reads attentively The cat quickly." Label: }

\textbf{A }

\textbf{B (Respond in one letter and nothing else)}
\end{tcolorbox}
\newpage

\noindent\texttt{\textbf{\textsc{H-Test}: End Ly}}
\begin{tcolorbox}[breakable, enhanced]

Input: "Neighbor ticks to the city skillfully." Label: A

Input: "He calculates all night high." Label: B

Input: "Dog sings with precision consistently." Label: A

Input: "Car travels in the kitchen hard." Label: B

Input: "Scientist calculates in the morning enthusiastically." Label: A

Input: "Bird cooks through the telescope very." Label: B

Input: "Book dances on the wall smoothly." Label: A

Input: "Dancer travels through the telescope high." Label: B

Input: "Dancer glows all night creatively." Label: A

Input: "River talks in the hall early." Label: B

Input: "Neighbor talks to the city creatively." Label: A

Input: "Teacher listens through the telescope quite." Label: B

Input: "Tree sings in the sky eagerly." Label: A

Input: "Car flows in space fast." Label: B

Input: "Tree sings through the telescope intently." Label: A

Input: "Tree sleeps in the hall far." Label: B

Input: "Dancer glows in space smoothly." Label: A

Input: "Friend travels in the forest fast." Label: B

Input: "Painter teaches with precision steadily." Label: A

Input: "Moon calculates in the morning so." Label: B

Input: "Cat flows in the studio eagerly." Label: A

Input: "Dog grows delicious meals too." Label: B

Input: "Cat reads in the hall carefully." Label: A

Input: "Scientist talks every second long." Label: B

Input: "Car cooks in the forest attentively." Label: A

Input: "Car listens in the hall wrong." Label: B

Input: "Chef jumps on the wall accurately." Label: A

Input: "Scientist reads delicious meals long." Label: B

Input: "Tree grows in space steadily." Label: A

Input: "Friend teaches with passion near." Label: B

Input: "Bird drives in the studio consistently." Label: A

Input: "Book teaches to music high." Label: B

Input: "Tree observes along the valley smoothly." Label: A

Input: "Dancer travels to the city too." Label: B

Input: "Neighbor travels with interest quietly." Label: A

Input: "Tree reads along the valley right." Label: B

Input: "Dancer glows every second melodiously." Label: A

Input: "Moon travels in the forest fast." Label: B

Input: "Tree glows delicious meals carefully." Label: A

Input: "Friend travels to music long." Label: B

Input: "Cat calculates in the sky brightly." Label: A

Input: "Bird listens along the valley enough." Label: B

Input: "Cat listens through the telescope steadily." Label: A

Input: "Clock glows in the studio quite." Label: B

\textbf{Input: "Tree calculates every second long." Label: }

\textbf{A }

\textbf{B (Respond in one letter and nothing else)}
\end{tcolorbox}
\newpage

\noindent\texttt{\textbf{\textsc{H-Test}: Palindrome}}
\begin{tcolorbox}[breakable, enhanced]

Input: "deed" Label: A

Input: "dwelt" Label: B

Input: "sis" Label: A

Input: "alight" Label: B

Input: "garage" Label: A

Input: "preses" Label: B

Input: "shahs" Label: A

Input: "phase" Label: B

Input: "ono" Label: A

Input: "strait" Label: B

Input: "Qaanaaq" Label: A

Input: "towd" Label: B

Input: "krk" Label: A

Input: "fe" Label: B

Input: "yay" Label: A

Input: "fumy" Label: B

Input: "Aviva" Label: A

Input: "kelt" Label: B

Input: "eve" Label: A

Input: "pien" Label: B

Input: "tat" Label: A

Input: "chaff" Label: B

Input: "heh" Label: A

Input: "Aymara" Label: B

Input: "alala" Label: A

Input: "xenia" Label: B

Input: "aeaea" Label: A

Input: "live" Label: B

Input: "Akasaka" Label: A

Input: "undid" Label: B

Input: "minim" Label: A

Input: "toluic" Label: B

Input: "adda" Label: A

Input: "sair" Label: B

Input: "dewed" Label: A

Input: "demal" Label: B

Input: "stots" Label: A

Input: "shun" Label: B

Input: "zerorez" Label: A

Input: "Skef" Label: B

Input: "peep" Label: A

Input: "eyrie" Label: B

Input: "sagas" Label: A

Input: "shee" Label: B

\textbf{Input: "tut" Label: }

\textbf{A }

\textbf{B (Respond in one letter and nothing else)}
\end{tcolorbox}
\newpage

\noindent\texttt{\textbf{\textsc{H-Test}: Spelled Math}}
\begin{tcolorbox}[breakable, enhanced]

Input: "The cube of approximately equal to 11." Label: A

Input: "The integral + 7." Label: B

Input: "The integral less than 11." Label: A

Input: "The integral / 7." Label: B

Input: "The integral times 10." Label: A

Input: "The sum + 10." Label: B

Input: "The square root plus 8." Label: A

Input: "The square root / 4." Label: B

Input: "The integral to the power of 12." Label: A

Input: "The cube of = 7." Label: B

Input: "The value of x times 12." Label: A

Input: "The sum > 8." Label: B

Input: "The square root approximately equal to 8." Label: A

Input: "The square root < 6." Label: B

Input: "The cube of equals to 6." Label: A

Input: "The sum + 10." Label: B

Input: "The fraction plus 9." Label: A

Input: "The cube of \^{} 8." Label: B

Input: "The product times 8." Label: A

Input: "The product > 11." Label: B

Input: "The derivative approximately equal to 5." Label: A

Input: "The limit \% 5." Label: B

Input: "The cube of plus 6." Label: A

Input: "The value of x < 6." Label: B

Input: "The value of x divided by 10." Label: A

Input: "The limit \% 9." Label: B

Input: "The integral approximately equal to 8." Label: A

Input: "The value of x * 9." Label: B

Input: "The derivative equals to 12." Label: A

Input: "The product \% 9." Label: B

Input: "The sum approximately equal to 6." Label: A

Input: "The value of x - 3." Label: B

Input: "The integral divided by 4." Label: A

Input: "The product < 12." Label: B

Input: "The limit to the power of 3." Label: A

Input: "The square root > 10." Label: B

Input: "The angle plus 3." Label: A

Input: "The product \% 12." Label: B

Input: "The square root to the power of 9." Label: A

Input: "The integral + 3." Label: B

Input: "The derivative times 9." Label: A

Input: "The sum \^{} 6." Label: B

Input: "The product plus 11." Label: A

Input: "The sum - 4." Label: B

\textbf{Input: "The limit < 10." Label: }

\textbf{A }

\textbf{B (Respond in one letter and nothing else)}
\end{tcolorbox}
\newpage

\noindent\texttt{\textbf{\textsc{H-Test}: Hyphenated Word}}
\begin{tcolorbox}[breakable, enhanced]

Input: "An eagle air-drops to the city quietly." Label: A

Input: "This engine reads to music fast." Label: B

Input: "An octopus part-times through the telescope expertly." Label: A

Input: "This engine teaches with passion intently." Label: B

Input: "This engine observes a double-check enthusiastically." Label: A

Input: "Our artist reads in the park attentively." Label: B

Input: "An octopus observes a part-time job expertly." Label: A

Input: "This engine jumps all night eagerly." Label: B

Input: "This engine proof-reads to music carefully." Label: A

Input: "The vendor listens delicious meals expertly." Label: B

Input: "An apple drives a mass-produced item slowly." Label: A

Input: "This engine drives to music intently." Label: B

Input: "An apple part-times through the telescope slowly." Label: A

Input: "My umbrella jumps to music quietly." Label: B

Input: "An iguana listens a long-term plan quietly." Label: A

Input: "An octopus listens with passion attentively." Label: B

Input: "Long-term talks quickly fast." Label: A

Input: "Our artist listens to the city enthusiastically." Label: B

Input: "The elephant runs a user-friendly interface enthusiastically." Label: A

Input: "An iguana sleeps to music attentively." Label: B

Input: "Part-time cooks with passion quietly." Label: A

Input: "This engine cooks through the telescope intently." Label: B

Input: "An eagle air-drops quickly expertly." Label: A

Input: "An eagle drives to music happily." Label: B

Input: "An apple air-drops to music slowly." Label: A

Input: "An iguana talks all night quietly." Label: B

Input: "An apple proof-reads quickly attentively." Label: A

Input: "The vendor runs all night slowly." Label: B

Input: "Double-check observes with passion slowly." Label: A

Input: "An iguana observes quickly expertly." Label: B

Input: "An octopus double-checks with passion happily." Label: A

Input: "An octopus cooks delicious meals slowly." Label: B

Input: "An owl teaches a user-friendly interface happily." Label: A

Input: "An iguana cooks to music quietly." Label: B

Input: "An eagle air-drops to music attentively." Label: A

Input: "An apple observes quickly expertly." Label: B

Input: "This engine sleeps a mother-in-law carefully." Label: A

Input: "An eagle reads with passion eagerly." Label: B

Input: "An iguana baby-sits in the park happily." Label: A

Input: "My umbrella reads to the city attentively." Label: B

Input: "Mother-in-law teaches to music quietly." Label: A

Input: "An owl talks in the park fast." Label: B

Input: "Well-being teaches on the wall eagerly." Label: A

Input: "This engine cooks all night happily." Label: B

\textbf{Input: "My umbrella observes with passion happily." Label: }

\textbf{A }

\textbf{B (Respond in one letter and nothing else)}
\end{tcolorbox}

\newpage

\section{Anecdotal Examples}
\label{App:anecdotal}
\begin{figure}[h]
\centering
\includegraphics[width=0.7\textwidth]{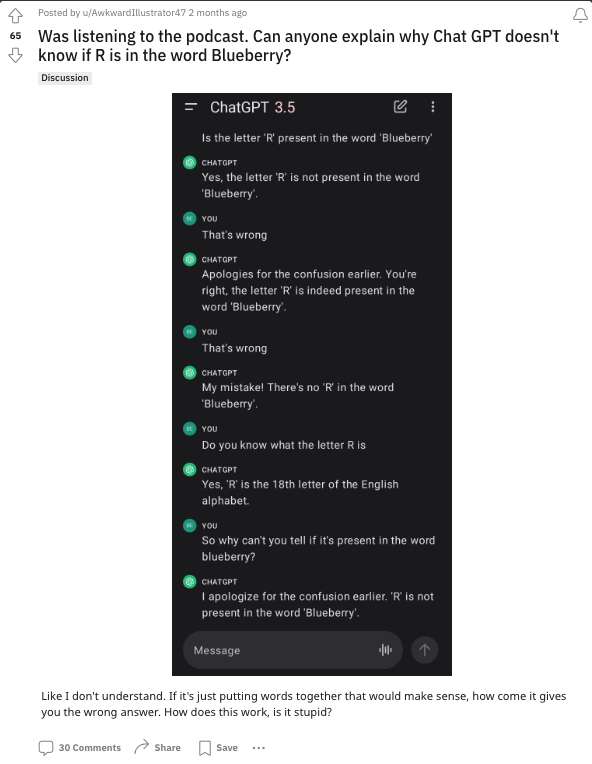}
\caption{\textbf{Blueberry?:} From Reddit u/AwkwardIllustrator47, r/mkbhd: \textit{Was listening to the podcast. Can anyone explain why Chat GPT doesn't know if R is in the word Blueberry?}}
\label{fig:blueberry}
\vspace{-4mm}
\end{figure}


\end{document}